\documentclass[10pt,twocolumn,letterpaper]{article}

\usepackage[pagenumbers]{cvpr}   %

\usepackage[utf8]{inputenc}
\usepackage[T1]{fontenc}
\usepackage{mathtools}
\usepackage{microtype}
\usepackage{array}
\usepackage{multirow}
\newcommand{\beh}{\mathcal{B}}

\newcommand{\argmax}{\operatorname*{arg\,max}}

\usepackage{float}
\usepackage{amssymb}

\definecolor{cvprblue}{rgb}{0.21,0.49,0.74}
\usepackage[pagebackref,breaklinks,colorlinks,allcolors=cvprblue]{hyperref}
\usepackage{orcidlink}

\def\confName{CVPR}
\def\confYear{2027}

\title{Detector Operating Behavior Inference via Data-Space Sampling}

\author{%
Stefan Becker$^{1}$\,\orcidlink{0000-0001-7367-2519} \quad
Simon Weiss$^{1}$\,\orcidlink{0009-0008-8171-9397} \quad
Wolfgang H\"ubner$^{1}$\,\orcidlink{0000-0001-5634-6324}\\
$^{1}$Fraunhofer IOSB, Ettlingen, Germany\\
Fraunhofer Center for Machine Learning\\
{\tt\small \{firstname.lastname\}@iosb.fraunhofer.de}
}

\begin{document}
\maketitle
\begin{abstract}
A detector scores candidate detections, but the recall and false alarms at a given confidence threshold depend on the input distribution. We introduce \emph{Detector Operating Behavior Inference} (DOBI), a second-level inference task: estimating a fixed detector's threshold-dependent, distribution-level operating response from the internal evidence of a single unlabeled image. DOBI treats the detector's threshold response, i.e.\ recall, false positives, and precision across thresholds, as a latent behavioral state under the current data condition. Controlled data-space sampling probes this state space offline, and a lightweight monitor learns the inverse map from one image to this state, whose single functional output supports multiple operating policies. In a controlled proof of concept with a fixed YOLOv10-Nano person detector and COCO-derived degradations, predicted PR behavior reduces mean absolute deviation from 0.128 for an image-independent mean-behavior prior to 0.054, and predicted functional distances track empirical regime distances (Spearman 0.96). From one unlabeled image, the inferred behavior selects operating points as well as a non-deployable lookup that knows the true regime (macro-regime $F_1$ 0.6604 for both, versus 0.6407 under pristine calibration; oracle 0.6620). It also supports validation-calibrated FPPI-constrained operation. On this discrete benchmark, regime-supervised curve lookups are strong controls. Pre-specified continuous and compositional probing shows where functional inference matters: on unseen combinations, the functional decoder has a lower mean curve error than a template lookup, increasingly with distance from the training conditions, whereas the lookup is more accurate for interpolation. These results establish the feasibility of single-image inference of distribution-level detector behavior over the probed conditions.\iftoggle{cvprfinal}{ Source code is available in our \href{https://github.com/beckerio/dobi}{code repository}.}{}
\end{abstract}

\section{Introduction}
\label{sec:introduction}

An object detector assigns scores to candidate boxes, and a confidence threshold decides which detections are reported. Yet the same threshold yields different recall, precision, and false-alarm rates when the image distribution changes---for example, under blur, noise, or fog. Confidence scores tell us which predictions survive a threshold, not what operating trade-off that threshold achieves under the current conditions.

Runtime monitoring has largely addressed two questions. \textbf{Q1: Is the input different from nominal data?} motivates shift detection. \textbf{Q2: Are the predictions reliable?} motivates uncertainty estimation, calibration, and failure prediction. DOBI asks a third: \textbf{Q3: How is the fixed detector operating under the current condition?} Its answer is a threshold-dependent functional response of the detector.

One way to explore different detector behaviors would be to vary model parameters. DOBI takes the complementary route: we keep one trained detector fixed and vary the data it sees. This sampling analogy motivates our experimental design, but does not imply that data changes reproduce arbitrary parameter changes. Each data condition induces an observable \emph{functional state}, described by TPR, FPR, FPPI, and precision across thresholds. Visually distinct conditions may induce similar responses, so we study the effect of a shift on the detector rather than its semantic label.

To learn these responses, we first need to measure them. Offline, we apply annotation-preserving transformations to labeled images, run the unchanged detector, and sweep its confidence threshold. The resulting curves supervise a monitor that operates without labels at inference. The selected transformations probe only part of the possible behavior space.

\begin{figure*}[t]
    \centering
    \includegraphics[width=\textwidth]{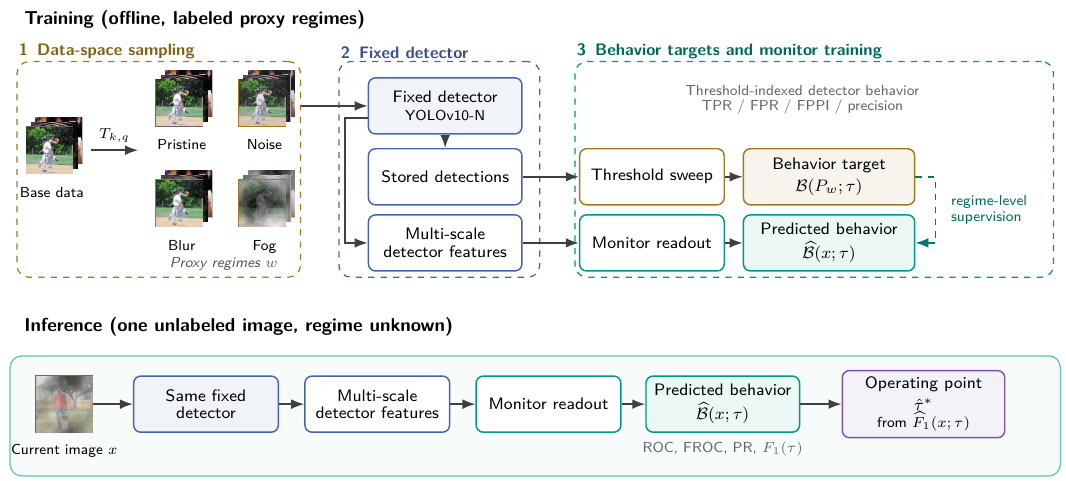}
    \caption{DOBI pipeline. \textbf{Offline:} controlled data-space probes expose regime-level behavior through threshold sweeps of an unchanged detector and supervise a feature-based monitor. \textbf{Online:} multi-scale features of one unlabeled image yield a threshold-indexed behavior estimate, supporting ROC/FROC/PR and operating policies without regime labels.}
    \label{fig:framework}
\end{figure*}

DOBI then addresses the \emph{inverse problem}. While the detector performs first-level inference of boxes, classes, and scores, a second-level monitor uses the detector's own internal response to infer how it operates under the latent condition. Regime identities only organize offline supervision; the monitor predicts a continuous response, without reconstructing the data distribution (\Cref{fig:framework}).

We keep the detector threshold $\tau$ as a common coordinate because ROC, FROC, PR, and operating points all arise from the same confidence sweep. The monitor predicts TPR, FPR, FPPI, and precision as functions of $\tau$, from which all operating curves and threshold policies follow.

We test whether probing reveals structured behavior (RQ1), whether single-image evidence recovers it (RQ2), whether the inferred response supports operating decisions (RQ3), whether regime-supervised lookups explain these results on a benchmark with discrete regimes (RQ4), and where functional inference outperforms template lookup when conditions are continuous or combined (RQ5).

Our contributions are:
\begin{itemize}
    \item We introduce \emph{detector operating behavior inference} as a second-level task: estimating the threshold-indexed functional state of a fixed detector, its ROC, FROC, and PR behavior under the current condition, from the internal evidence of a single unlabeled image.
    \item We use controlled \emph{data-space sampling as behavioral probing}: interventions act as experimental queries of a detector-specific operating-behavior space, in which conditions are similar when they induce similar responses.
    \item We solve the resulting inverse problem with a shared threshold-indexed behavior decoder. In a controlled proof of concept, single-image inference reconstructs regime-level curves and selects operating points as well as a lookup that knows the true regime.
    \item We provide diagnostic controls that separate behavior inference from regime lookup, scalar priors, decoder representation, and operating-point-only adaptation, and a pre-specified continuous and compositional test that shows where functional inference outperforms template lookup and where it does not.
\end{itemize}

\section{Related Work}
\label{sec:related_work}

The relevant literature differs mainly in the quantity it infers. Shift and image-quality methods characterize the input, uncertainty and calibration characterize current predictions, and runtime performance estimation predicts aggregate performance. DOBI targets a further object: the \emph{threshold-dependent operating response} of a fixed detector under the current latent condition.

\paragraph{Data shift, corruptions, and image quality.}
Common-corruption benchmarks established that controlled interventions in image space induce systematic performance changes \cite{hendrycks_ICLR_2019,michaelis_NeurIPSW_2019}. Their endpoint is robustness measurement after labels are available; we instead use controlled transformations as experimental probes that expose detector-specific functional states and supply supervision for an inverse behavior monitor. No-reference IQA methods such as BRISQUE, NIQE, and ARNIQA infer degradation or perceptual quality from an image \cite{mittal_TIP_2012,mittal_SPL_2013,agnolucci_WACV_2024}. This demonstrates that degradation information is locally observable, but perceptual quality is not detector behavior: visually similar shifts can affect detectors differently, and detector-relevant changes need not align with perceptual quality. Detector-aligned degradation representations narrow this gap by encoding fidelity directly in task features \cite{becker_ACCV_2026}.

\paragraph{Novelty, uncertainty, and reliability.}
Classical OoD methods ask whether an observation is atypical with respect to nominal data, using signals such as softmax confidence, ODIN, energy, or feature-space outliers \cite{hendrycks_ICLR_2017,liang_ICLR_2018,liu_NeurIPS_2020,du_ICLR_2022}. In object detection, novelty is especially different from operational degradation because dense structured outputs entangle semantics, localization, and confidence. Predictive uncertainty and calibration instead characterize the reliability of detections or prediction sets \cite{miller_CVPRW_2019,hall_WACV_2020,kuppers_CVPRW_2020,pathiraja_CVPR_2023}. Self-aware detection work has emphasized that robustness, uncertainty, calibration, and scene-level acceptance should be considered jointly \cite{oksuz_CVPR_2023}. These methods answer whether observations or predictions should be trusted, not what threshold response the detector currently induces.

\paragraph{Label-free performance estimation.}
The closest line of work estimates performance without deployment-time labels. For classification, AutoEval regresses accuracy from feature statistics of unlabeled test sets, trained on a meta-dataset of synthetically transformed sets \cite{deng_CVPR_2021}, and ATC predicts accuracy from thresholded model confidence \cite{garg_ICLR_2022}. For detection, perturbation sensitivity has been used for label-free robustness estimation \cite{shekar_IJCV_2021}; ROMA predicts runtime accuracy for detector selection \cite{lee_WACV_2023}; and automated evaluation estimates detection performance from prediction consistency and reliability cues \cite{yoo_ICCV_2025}. DOBI shares the idea of supervising a performance predictor with synthetically probed conditions, but differs in target and evidence: it infers a threshold-indexed function, from which ROC/FROC/PR behavior and several operating policies follow, from a single image rather than a scalar from an image set. Because the output targets differ, our baselines test the functional-inference hypothesis directly rather than re-purposing scalar estimators.

\paragraph{Discrete condition representations and functional behavior.}
Condition recognition provides one natural way to organize deployment states: a finite set of named conditions can be associated with precomputed performance summaries or operating points. DOBI takes a complementary, more functional view. The proxy conditions index samples of the detector-specific mapping from data distributions to behavior, while the monitor predicts the induced threshold response itself. A discrete condition model can therefore be understood as a coarse representation of this behavior space in which each label selects one stored template. Direct functional prediction requires no named condition at inference and can represent similarities between sampled states.

\section{Method}
\label{sec:method}

Let $f_{\hat{\theta}}$ be a fixed detector with unchanged parameters and let $z_{\hat{\theta}}(x)$ denote the input-dependent multi-scale internal evidence it produces for image $x$. The detector performs first-level task inference,
\begin{equation}
    f_{\hat{\theta}}(x)=\{(b_j,c_j,s_j)\}_{j=1}^{N_x},
\end{equation}
where $b_j$, $c_j$, and $s_j$ are a box, class, and score. DOBI adds a second behavioral inference layer,
\begin{equation}
    g_{\phi}\!\left(z_{\hat{\theta}}(x),\tau\right)
    =\widehat{\beh}_{\hat{\theta}}(x;\tau),
\end{equation}
which estimates the fixed detector's threshold response.

\subsection{Detector-Specific Functional State Space}
\label{sec:behavior_space}
For a latent operating condition $w$ with image distribution $P_w$ and detector confidence threshold $\tau$, we describe the fixed detector by
\begin{equation}
\begin{aligned}
\beh_{\hat{\theta}}(P_w;\tau)=\bigl(&\mathrm{TPR}_w(\tau),\mathrm{FPR}_w(\tau),\\
&\mathrm{FPPI}_w(\tau),\mathrm{Prec}_w(\tau)\bigr).
\end{aligned}
\label{eq:behavior}
\end{equation}
ROC, FROC, and PR are projections of this same threshold-indexed object. Under the adopted matching protocol, recall equals TPR; FROC and PR use detection-native FPPI and precision, while the candidate-normalized $\mathrm{FPR}(\tau)=\mathrm{FP}(\tau)/\mathrm{FP}(\tau_0)$ divides false positives by the unmatched candidates at the score floor $\tau_0=0.001$.

The fixed detector thus induces a map $\Gamma_{\hat\theta}: P\mapsto\beh_{\hat\theta}(P;\cdot)$ from data conditions to functional operating behavior. Its image $\mathcal{S}_{\hat\theta}=\{\Gamma_{\hat\theta}(P)\mid P\in\mathcal{P}\}$ forms the detector's \emph{effective functional state space}: observable operating responses under varying distributions, not a mutable network state.

Within this space, two data conditions are \emph{behaviorally similar} at tolerance $\varepsilon$ when the distance $d_{\beh}$ between their responses is at most $\varepsilon$. Distinct conditions may thus induce nearly identical responses; for a retrained detector this relation may differ. The target is detector-specific behavior rather than data-distribution reconstruction.

\subsection{Controlled Data-Space Probing}
We probe $\Gamma_{\hat\theta}$ with controlled interventions. For a reference distribution $P_0$ and transformation $T_{k,q}$ (type $k$, severity $q$), we construct
\begin{equation}
    x_{k,q}=T_{k,q}(x)\sim P_{k,q},
    \qquad
    \beh_{k,q}=\Gamma_{\hat\theta}(P_{k,q}),
\end{equation}
where $P_{k,q}$ denotes the transformed distribution and detector sweeps reveal its behavior $\beh_{k,q}$.
The pairs $\{(P_m,\Gamma_{\hat\theta}(P_m))\}_{m=1}^{M}$ \emph{partially chart} the state space through repeatable, measurable effects rather than a fixed ontology of degradation labels.

Varying model parameters offers a conceptual analogy for generating diverse operating responses, not a hypothesis tested by DOBI. At fixed $P_0$, parameter interventions would sample $\{\beh_{\theta}(P_0;\cdot):\theta\in\Theta\}$, whereas DOBI samples $\{\beh_{\hat\theta}(P;\cdot):P\in\mathcal{P}_{\mathrm{probe}}\}$ with $\hat\theta$ fixed. We neither assume nor test equivalence between these families; only measurable data-induced variation under the selected probes is required.

The main study selects degradations and severities deliberately, rather than using random Monte-Carlo sampling. This controlled identifiability study tests whether local evidence recovers states from a reproducibly sampled portion of $\mathcal{S}_{\hat\theta}$; RQ5 adds randomly drawn continuous strengths and pairwise combinations (\Cref{sec:continuous}).

\subsection{Empirical Supervision and Inverse Behavioral Inference}
\label{sec:inverse}
The fixed detector is evaluated offline on every proxy regime with identical inference settings. Stored predictions and annotations are swept over detector score thresholds to obtain empirical tuples $(\tau_i,\mathrm{TPR}_i,\mathrm{FPR}_i,\mathrm{FPPI}_i,\mathrm{Prec}_i)$. Each image inherits its regime-level behavior target. IDs associate training samples with targets but are never monitor inputs or predictions.

For an image $x\sim P_w$, the multi-scale feature maps produced by the fixed detector are pooled and passed through a lightweight readout to obtain $z_{\hat{\theta}}(x)$. The monitor learns the inverse direction of the sampled forward map,
\begin{equation}
\widehat{\beh}(x;\tau)\approx
\mathbb{E}\!\left[\Gamma_{\hat\theta}(P_w)(\tau)\mid z_{\hat\theta}(x)\right].
\label{eq:inverse}
\end{equation}
Under a squared loss, to which our Smooth-$L_1$ loss reduces for small residuals, \Cref{eq:inverse} is the optimal predictor; a single image only needs to carry evidence about the condition, not to reconstruct $P_w$. For discrete probing regimes $w$ with curves $\beh_w$, it is the posterior mixture $\sum_w p(w\mid z_{\hat\theta}(x))\,\beh_w(\tau)$: it reduces to the mean-behavior prior when the evidence carries no information about $w$, and to the regime's curve when $w$ is identified. The probing regimes need no semantic meaning; they only define where detector behavior is measured. This finite-mixture form follows from the discrete probing design; under continuous or compositional probing (RQ5), the posterior ranges over behavior functions. What matters is uncertainty over behavior functions rather than condition names; confusing two regimes with similar curves costs little.

The detector remains fixed throughout; retraining it changes $\Gamma_{\hat\theta}$ and the targets, not the DOBI pipeline.

\subsection{Threshold-Indexed Behavior Decoder}
We use $K$ global knots $0\leq\tau_1<\cdots<\tau_K\leq1$. The monitor predicts TPR, FPR, FPPI, and precision at these knots. TPR, FPR, and FPPI are constrained to be non-increasing as $\tau$ rises by parameterizing positive decrements; TPR/FPR are bounded to $[0,1]$, FPPI is modeled in $\log(1+\mathrm{FPPI})$ space, and precision is bounded but not forced to be monotonic. A shape-preserving piecewise cubic Hermite interpolant (PCHIP) evaluates the functions between knots; piecewise-linear interpolation is the main representation ablation.

From the predicted functions we derive
\begin{equation}
\widehat F_1(x;\tau)=
\frac{2\,\widehat{\mathrm{Prec}}(x;\tau)\widehat{\mathrm{TPR}}(x;\tau)}
{\widehat{\mathrm{Prec}}(x;\tau)+\widehat{\mathrm{TPR}}(x;\tau)+\epsilon},
\end{equation}
and use $\widehat\tau^{*}(x)=\argmax_{\tau}\widehat F_1(x;\tau)$ as one concrete operating policy. This operating point is a downstream readout of the inferred function; the same output also supports other policies, such as maximizing TPR under an FPPI constraint, without retraining the monitor.

The main model is trained on 201 threshold samples with equally weighted Smooth-$L_1$ losses for TPR, FPR, FPPI, and precision ($\beta=0.05$). No direct threshold loss or regime-classification loss is used in the main model.

\section{Experimental Setup}
\label{sec:experiments}

We evaluate five questions: \textbf{RQ1} whether controlled data-space sampling exposes structured detector behavior, \textbf{RQ2} whether a second-level monitor can infer that regime-level behavior from local detector evidence, \textbf{RQ3} whether the inferred function provides an operational benefit over pristine calibration, \textbf{RQ4} whether regime-supervised lookups explain these results (\Cref{sec:followup}), and \textbf{RQ5} where functional inference outperforms template lookup under continuous and compositional probing (\Cref{sec:continuous}).

\paragraph{Dataset and proxy regimes.}
We use COCO~\cite{lin_ECCV_2014} for person detection. \emph{COCOMini is used only to construct the monitor training pool.} After person-focused filtering, this pool contains 8,706 base images; 870 are held out for monitor validation and 7,836 remain for training. The independent test pool contains 1,650 person-focused images from the regular COCO validation side, not from COCOMini. We apply 17 degradations (noise, blur, weather, color/contrast, compression, and pixelation families). The main monitor groups severities 3--5 by degradation type, yielding 18 proxy regimes: pristine plus 17 degraded regimes. Each base image therefore has 52 stored variants, giving 407,472 training, 45,240 validation, and 85,800 test variants. Severities 1--5 are used for the separate severity analysis.

\paragraph{Detector and monitor.}
The fixed base detector is YOLOv10-Nano~\cite{wang_NeurIPS_2024} at $640\times640$ resolution; only person detections are analyzed. Matching is greedy in descending score order at IoU 0.5, stored detections extend to score 0.001, and FROC uses $\mathrm{FPPI}\in[0,2]$. The monitor pools three FPN levels with global average and max pooling and uses a two-layer MLP. The main model uses $K=16$ uniform threshold knots, PCHIP interpolation, and $\log(1+\mathrm{FPPI})$ prediction. Models are trained for 60 epochs with AdamW ($10^{-4}$), cosine scheduling, batch size 64, mixed precision, and three seeds; the best validation-loss checkpoint is selected.

\paragraph{Baselines and metrics.}
The \emph{fixed pristine} baseline uses one pristine-validation threshold ($\tau=0.2643$). The image-independent \emph{mean-behavior prior} averages the four threshold functions over training regimes and derives one global operating point ($\tau=0.175$). We further compare linear interpolation and a one-dimensional PC1 behavior model. The non-deployable \emph{Regime-ID} baseline is a discrete-condition diagnostic: it assumes perfect knowledge of the constructed regime and retrieves that regime's validation-optimal threshold. A test-label regime oracle provides a non-deployable reference that uses the test-optimal fixed threshold of each regime.

Curve reconstruction is measured by mean absolute deviation (MAD) on the established ROC, FROC, and PR grids. Operational evaluation uses all 85,800 test variants: each predicted threshold is applied only to detections of its own image, then TP/FP/FN are aggregated. We report pooled micro metrics and macro averages over the 18 regimes. Macro $F_1$ is the mean of detector $F_1$ computed separately per regime; it is primary because the question concerns consistent operation across conditions. Regret is the macro $F_1$ gap to the regime oracle. Paired 95\% bootstrap intervals resample the 1,650 base images 2,000 times while retaining their corresponding variants. Ablations and further diagnostics are reported in the supplement.

\section{Results}
\label{sec:results}

\subsection{RQ1: Do Data-Space Probes Reveal a Structured Functional State Space?}
\label{sec:behavior_results}
\begin{figure*}[t]
    \centering
    \includegraphics[width=\textwidth]{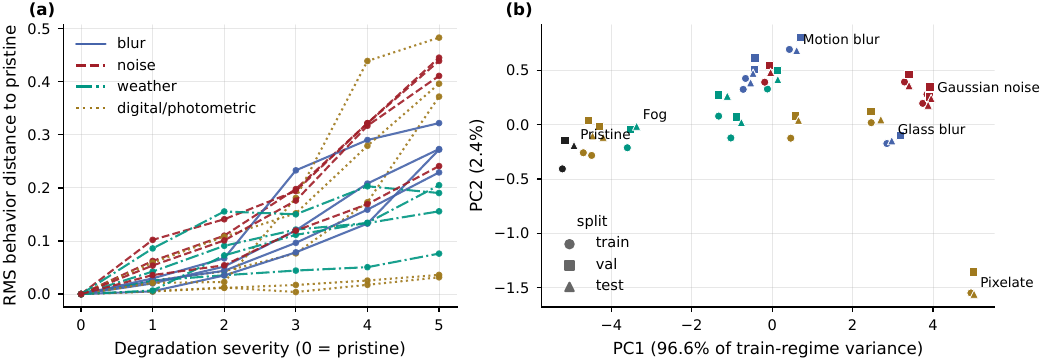}
    \caption{Behavior exposed by controlled data-space probes. (a) RMS distance from pristine versus degradation severity. (b) Train/validation/test regime behavior projected onto PCA fitted on training regimes only. PC1 explains 96.6\% for the \emph{shown, specifically weighted behavior vector}; other component weightings reveal additional structure (supplement).}
    \label{fig:data_space}
\end{figure*}

The first question is whether varying data conditions reveals repeatable changes in the detector's operating curves. The distance from pristine behavior increases with severity for 15 of 17 degradation types (\Cref{fig:data_space}a), indicating that intervention strength produces systematic motion through behavior space rather than arbitrary curve variation. The same regime differs between train, validation, and test by only 0.009 RMS on average, whereas different regimes differ by 0.13 on average. For the specific behavior vector and component weighting in \Cref{fig:data_space}b, PCA fitted on training regimes assigns 96.6\% of variance to PC1 and 2.4\% to PC2. This is representation-dependent, not an intrinsic dimension of detector behavior (supplement). The $F_1$-optimal threshold varies systematically along this structure, from about 0.27--0.28 for pristine-like states to about 0.05 for the Pixelate state. Together, these observations show that the proxy conditions provide repeatable samples of a structured detector-specific state space.

\begin{figure}[t]
    \centering
    \includegraphics[width=\columnwidth]{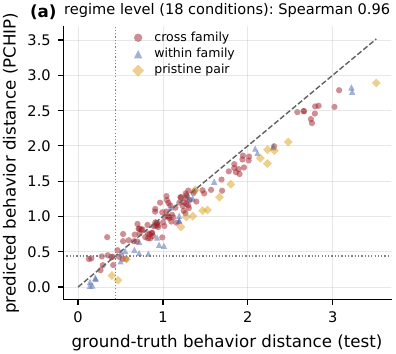}
    \caption{Functional state-space geometry. Each point is a pair of regimes; colors distinguish cross-family, within-family, and pristine pairs. Empirical versus PCHIP-predicted distances use a standardized behavior metric (Spearman $0.96$); the dashed line denotes equality. This tests recovery of within-detector distance geometry, not data/parameter intervention equivalence.}
    \label{fig:equivalence}
\end{figure}

\paragraph{Functional similarity across data conditions.}
A separate cross-family analysis measures approximate behavioral similarity between data conditions. Using a validation-derived measurement tolerance and a standardized threshold-domain behavior distance, conditions from different corruption families can exhibit approximately matching responses on disjoint test images. Corruption-family membership only weakly predicts functional closeness (AUROC about 0.55). More importantly, predicted pairwise behavior distances agree with empirical distances across the 18 regimes (Spearman $0.96$; \Cref{fig:equivalence}). Agreement falls to $0.75$ across 52 severity-resolved conditions. Matches depend on distance and tolerance (supplement).

\subsection{RQ2: Can Local Evidence Recover Distribution-Level Behavior?}
This is the primary inverse-inference test of DOBI. The monitor receives only local multi-scale detector evidence but must reconstruct a distribution-level functional target. Compared with the mean-behavior prior, PCHIP reduces ROC MAD from 0.108 to 0.043, FROC MAD from 0.129 to 0.034, and PR MAD from 0.128 to 0.054 (\Cref{tab:curvefit}). Linear interpolation is close, indicating that preserving $\tau$ as a common coordinate matters more than the interpolation family itself. In a resampling analysis, aggregating independent images from one regime lowers PCHIP PR MAD from $0.0534$ (one image) to $0.0439$ (32 images), approaching the $0.0435$ all-image limit. Additional evidence thus reduces sampling variation, but not the remaining systematic regime-level bias (supplement). A detector-specific replication on YOLOv10-m and YOLO11-m likewise reduces ROC/FROC MAD relative to the mean-behavior prior by 76--83\%/81--85\% with PCHIP; full results and limitations appear in the supplement.

\begin{table}[t]
\centering
\caption{Curve reconstruction on the 18 test regimes (macro average). Learned models: mean $\pm$ standard deviation over three seeds.}
\label{tab:curvefit}
\resizebox{\columnwidth}{!}{%
\begin{tabular}{lccc}
\toprule
Model & ROC MAD & FROC MAD & PR MAD \\
\midrule
Mean behavior & 0.1077 & 0.1287 & 0.1278 \\
Linear & $0.0468\!\pm\!0.0013$ & $\mathbf{0.0339}\!\pm\!0.0004$ & $0.0587\!\pm\!0.0007$ \\
Adaptive PCHIP & $\mathbf{0.0433}\!\pm\!0.0015$ & $0.0342\!\pm\!0.0005$ & $\mathbf{0.0536}\!\pm\!0.0006$ \\
\bottomrule
\end{tabular}
}
\end{table}

\begin{figure*}[t]
    \centering
    \includegraphics[width=\textwidth]{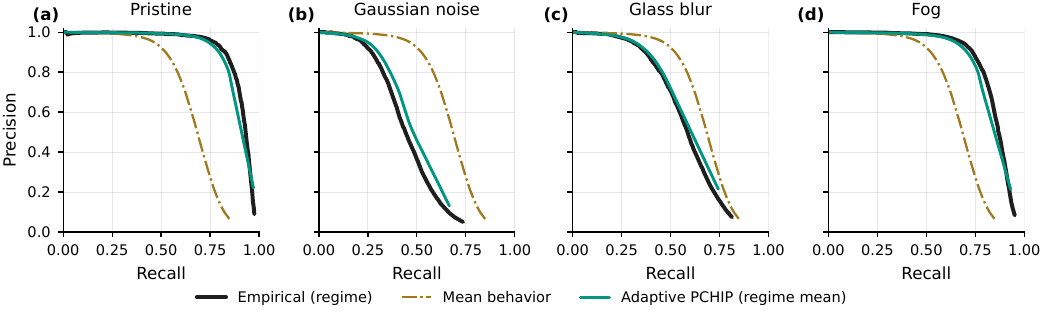}
    \caption{Precision--Recall behavior for representative test regimes. Black: empirical regime target. Green: average image-conditioned PCHIP prediction over test images and three seeds. Dash-dotted: mean-behavior prior.}
    \label{fig:curvefit}
\end{figure*}

\Cref{fig:curvefit} emphasizes PR because it is detection-native and directly exposes the precision--recall trade-off. The average image-conditioned prediction tracks the regime-specific target under both severe shifts (Gaussian noise, Glass blur) and near-pristine conditions, whereas the mean-behavior prior has one fixed shape. Full ROC/FROC and threshold-domain functions, restricted to their empirically realized segments, are in the supplement.

\subsection{RQ3: Does the Inferred Function Provide a Concrete Benefit?}
As a downstream utility test, we ask whether the reconstructed function contains enough information to support an operating decision. Under true per-image execution, the pristine-calibrated threshold obtains macro-regime $F_1=0.6407$ (\Cref{tab:operation}). A single global mean-behavior threshold already improves this to 0.6536, showing that pristine calibration is substantially mismatched to the degraded regime mixture. Adaptive PCHIP reaches 0.6604, close to the regime oracle at 0.6620. Its gain over pristine calibration is $+0.0197$ with paired 95\% bootstrap interval $[0.0176,0.0217]$, recovering about 93\% of the pristine-to-oracle gap. Relative to the strong mean-behavior prior, PCHIP reduces macro execution regret from 0.0083 to 0.0016.

\begin{table}[t]
\centering
\caption{True threshold execution on all 85,800 test variants. Macro $F_1$ is the mean of detector $F_1$ across regimes, not regime-classification $F_1$; micro $F_1$ pools detections. Regret is the macro gap to the regime oracle. $^\dagger$Uses regime identity and is non-deployable.}
\label{tab:operation}
\resizebox{\columnwidth}{!}{%
\begin{tabular}{lcccc}
\toprule
Policy & Macro $F_1$ & Micro $F_1$ & Regret & FPPI \\
\midrule
Fixed pristine & 0.6407 & 0.6494 & 0.0212 & 0.148 \\
Mean behavior & 0.6536 & 0.6617 & 0.0083 & 0.285 \\
PC1 behavior & $0.6601\!\pm\!0.0001$ & 0.6615 & 0.0018 & 0.339 \\
Adaptive PCHIP & $0.6604\!\pm\!0.0002$ & 0.6622 & 0.0016 & 0.327 \\
\midrule
Regime-ID$^\dagger$ & 0.6604 & 0.6605 & 0.0016 & 0.344 \\
Regime oracle$^\dagger$ & 0.6620 & 0.6625 & 0.0000 & 0.324 \\
\bottomrule
\end{tabular}
}
\end{table}

For pooled micro $F_1$, PCHIP improves over pristine by $+0.0129$ but exceeds the mean-behavior prior by only $+0.0005$ (CI includes zero). Thus the clearer gain is \emph{consistency across regimes}.

Per regime (supplement), PCHIP improves 14 of 18 regimes, with the largest gains for Glass blur ($+0.068$), Pixelate ($+0.056$), and Gaussian noise ($+0.043$). The mechanism is recall recovery: adaptation lowers the threshold, increasing pooled recall from 0.512 to 0.566 while precision decreases from 0.888 to 0.799 and FPPI rises from 0.148 to 0.327. This trade-off follows from optimizing $F_1$; adaptation does not reduce false alarms here.

\paragraph{A second operating objective.}
The same predicted functions also support maximizing recall subject to an FPPI budget. With $B=0.1$, a validation-calibrated PCHIP policy reaches macro recall $0.462$ versus $0.438$ for a conservative fixed threshold, with no observed budget exceedance across the 18 test regimes (supplement). Without calibration, predicted FPPI is not a reliable constraint: uncalibrated PCHIP violates the budget in an average of 10.3 regimes. Calibration uses validation labels only, and the absence of test violations is not a guarantee under unseen shifts. The $F_1$ and FPPI policies require different operating decisions, but both derive from the same inferred behavior function.

\subsection{What Does the Second-Level Monitor Recover?}
The non-deployable Regime-ID baseline knows the true regime and applies its validation-optimal threshold. From a single unlabeled image, PCHIP reaches the same macro $F_1$ (0.6604); deployable regime-supervised lookups are within 0.001 (\Cref{sec:followup}). The monitor thus recovers the operationally relevant regime-level information without regime labels. Consistently, one fixed predicted threshold per regime performs as well as image-specific thresholds (0.6607 versus 0.6604): the monitor infers a regime-level functional state, not an intrinsic image-specific curve. A one-coefficient PC1 monitor nearly matches PCHIP for $F_1$ (0.6601), but not every FPPI policy (supplement).

\subsection{RQ4: Do Regime-Supervised Lookups Explain the Results?}
\label{sec:followup}
Because the present benchmark assigns images to 18 discrete proxy targets, regime-supervised classifiers are a strong diagnostic of how much of the learned behavior finite-template selection explains. We test these finite-template controls on the same detector, features, and test pool (supplement). Two \emph{regime-supervised controls} return the stored curve of the predicted training condition (hard lookup) or a probability-weighted mixture of stored curves (soft lookup); because they are trained on regime labels, they are controls rather than DOBI realizations. Two \emph{regime-label-free} variants change only the decoder: PCHIP with 32 uniform knots and a template decoder that mixes stored training curves with weights learned from the curve loss alone.

\begin{table}[t]
\centering
\caption{RQ4 follow-up studies on the same test pool. MAD: curve error over ROC, FROC, and PR (lower is better); $F_1$: macro $F_1$ under true execution. Sev.\,4 and sev.\,2+4: severities not seen in training (trained on $\{3,5\}$ and, as a pre-specified test, on $\{1,3,5\}$). The template decoder was not part of the pre-specified sev.\,2+4 test. $^\dagger$Trained with regime labels.}
\label{tab:followup}
\resizebox{\columnwidth}{!}{%
\begin{tabular}{lcccccc}
\toprule
 & Regime & \multicolumn{2}{c}{18 regimes} & sev.\,4 & sev.\,2+4 \\
\cmidrule(lr){3-4}\cmidrule(lr){5-5}\cmidrule(lr){6-6}
Arm & labels & MAD & $F_1$ & MAD & MAD \\
\midrule
Mean behavior & -- & 0.1214 & 0.6536 & 0.1275 & 0.1096 \\
PCHIP, 16 knots & no & 0.0437 & 0.6604 & 0.0850 & 0.0597 \\
PCHIP, 32 knots & no & 0.0362 & 0.6596 & 0.0783 & 0.0557 \\
Template decoder & no & 0.0291 & 0.6605 & 0.0846 & -- \\
Hard lookup$^\dagger$ & yes & 0.0245 & 0.6596 & 0.0988 & 0.0705 \\
Soft lookup$^\dagger$ & yes & 0.0256 & 0.6608 & 0.0795 & 0.0549 \\
\bottomrule
\end{tabular}
}
\end{table}

\Cref{tab:followup} gives three answers. First, on the 18 regimes, the lookups fit the curves best, as the posterior-mixture form (\Cref{sec:inverse}) anticipates; the classifier is correct for only 76\% of images but confuses regimes with similar curves. Second, regime-label-free decoders close much of this gap, and decisions are unaffected (macro $F_1$ within 0.002 across all learned arms). Third, a pre-specified test of whether the continuous decoder interpolates unseen severities better than the soft lookup was \emph{refuted} ($\Delta$ mean MAD $+0.0008$, less than the lookup's seed SD). Regime-level behavior is thus recoverable without regime labels, but this discrete benchmark shows no advantage of continuous over discrete representations; RQ5 tests where one arises.

\subsection{RQ5: Continuous and Compositional Probing}
\label{sec:continuous}
We sample 15 degradation types at continuous strengths and in pairwise combinations, each condition with its own target curve on 1,200 images. Monitors are trained on 307 conditions, whose stored curves form the soft lookup's bank. By \Cref{sec:inverse}, the lookup should be near-optimal between densely stored conditions, while a functional decoder can help where no stored condition is close. An initial experiment on 142 unseen conditions motivated three pre-specified hypotheses, tested with the frozen monitors on 710 fresh conditions (supplement).

\begin{table}[t]
\centering
\caption{RQ5, 710 fresh test conditions: functional decoder (32 knots) minus soft lookup. $\Delta$: mean curve-error difference (negative favors the decoder), basic 95\% CI; Median: median $\Delta_c$; Better: share of conditions the decoder wins. $^{1,2}$Pre-specified hypotheses.}
\label{tab:continuous}
\footnotesize
\setlength{\tabcolsep}{2pt}
\begin{tabular}{lrcccc}
\toprule
Test set & $n$ & $\Delta$ & 95\% CI & Median & Better \\
\midrule
Compositions$^{1}$ & 650 & $-0.0043$ & [$-0.0049$, $-0.0037$] & $+0.0013$ & 46\% \\
\quad unseen pairs & 416 & $-0.0063$ & [$-0.0071$, $-0.0055$] & $+0.0003$ & 49\% \\
\quad seen pairs & 234 & $-0.0008$ & [$-0.0012$, $-0.0002$] & $+0.0034$ & 41\% \\
Interpolation$^{2}$ & 60 & $+0.0067$ & [$+0.0065$, $+0.0076$] & $+0.0055$ & 2\% \\
\bottomrule
\end{tabular}

\end{table}

\begin{figure}[t]
\centering
\includegraphics[width=\columnwidth]{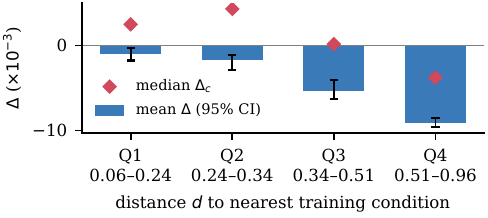}
\caption{Decoder advantage by distance to the nearest training condition (650 compositions, quartiles of $d$). Bars: mean $\Delta$ with basic 95\% CI; diamonds: median $\Delta_c$. Negative values favor the functional decoder.}
\label{fig:continuous}
\end{figure}

All three hypotheses are supported (\Cref{tab:continuous,fig:continuous}). On 650 unseen compositions, the functional decoder (PCHIP with 32 knots, as in RQ4) has a lower mean curve error than the soft lookup; on interpolation between stored strengths, the lookup is better; and the decoder's advantage grows with the distance $d$ to the nearest training condition (Spearman $\rho=-0.18$ [$-0.26$, $-0.10$]). The gain is a mean effect: the decoder wins only 46\% of compositions but exceeds the lookup's 90th-percentile validation error less often (13\% versus 18\%), mostly on type pairs never combined in training. It requires 32 knots: the 16-knot decoder of the main experiments is worse than the lookup on these compositions.

\paragraph{Limits of the main study.} Fog remains a small failure case ($-0.0039$ gain). In curve MAD, the main monitor is worse than the mean-behavior prior for 4 of 18 regimes, and learned monitors can fall below it outside charted support, where inference is least reliable (supplement).

\section{Discussion and Limitations}
\label{sec:discussion}

Controlled interventions expose functional detector states, and local multi-scale features suffice to infer them. Curve reconstruction is the primary evidence; operating-point selection and validation-calibrated FPPI control are downstream uses. Inference itself has a large effect, more than halving curve error against an image-independent prior, whereas differences between good realizations (functional decoders, lookups) are small and mark the open design space. Different conditions can induce similar responses for a fixed detector, although similarity depends on the chosen behavior metric.

\paragraph{Scope of the benchmark.} With 18 pooled regimes as discrete targets, regime-supervised lookups are especially strong controls, as RQ4 confirms, and regime-label-free decoders approach them. Continuous and compositional probing (RQ5) shows where functional inference pays off: on combinations far from stored conditions, mainly through fewer large errors, whereas lookups remain better between densely stored conditions. DOBI contributes single-image inference of distribution-level functional responses; state-space dimension and image-specific curves remain open.

\paragraph{Coverage and generality.} The parameter-space analogy motivated data-space probing, but is neither an assumption nor an empirical claim of DOBI. The principal limitation is functional-state coverage: synthetic degradations cannot represent every deployment shift. Richer probes could sample sensors, geography, or higher-order mixtures of conditions. Finally, the study centers on YOLOv10-Nano, one class, and a COCO-derived setting. Replications with YOLOv10-m and YOLO11-m reproduce the curve-inference gains but reveal a low-threshold PCHIP failure for YOLO11-m Pixelate; they do not establish transfer to other tasks or uncharted functional states. On BDD100K~\cite{yu_CVPR_2020}, monitors trained on natural scene and time-of-day regimes (chosen on vehicle data) beat the mean-behavior prior and a brightness baseline for vehicles and persons, with small operational gains and within one domain (supplement).

\section{Conclusion}
\label{sec:conclusion}

We introduced Detector Operating Behavior Inference as a second-level task: estimating the threshold-dependent, distribution-level behavior of a fixed detector from local multi-scale evidence. Controlled data-space interventions expose measurable functional states that group conditions by their effect on detector operation rather than appearance. Across the probed conditions, DOBI reconstructs behavior more accurately than an image-independent mean-behavior prior and selects operating points as well as a lookup that knows the true regime. Continuous and compositional probing shows where functional inference matters: on unseen combinations, it makes fewer large errors than a template lookup. These findings establish the feasibility of functional behavioral inference, while broader deployment shifts remain an open generalization challenge.

{\small
\bibliographystyle{ieeenat_fullname}
\bibliography{references}

\begin{thebibliography}{23}
\providecommand{\natexlab}[1]{#1}
\providecommand{\url}[1]{\texttt{#1}}
\expandafter\ifx\csname urlstyle\endcsname\relax
  \providecommand{\doi}[1]{doi: #1}\else
  \providecommand{\doi}{doi: \begingroup \urlstyle{rm}\Url}\fi

\bibitem[Agnolucci et~al.(2024)Agnolucci, Galteri, Bertini, and
  Bimbo]{agnolucci_WACV_2024}
Lorenzo Agnolucci, Leonardo Galteri, Marco Bertini, and Alberto~Del Bimbo.
\newblock {ARNIQA}: Learning distortion manifold for image quality assessment.
\newblock In \emph{IEEE/CVF Winter Conference on Applications of Computer
  Vision (WACV)}, pages 188--197, 2024.

\bibitem[Becker et~al.(2026)Becker, Weiss, H{\"u}bner, and
  Arens]{becker_ACCV_2026}
Stefan Becker, Simon Weiss, Wolfgang H{\"u}bner, and Michael Arens.
\newblock Self-aware object detection via degradation manifolds.
\newblock In \emph{Asian Conference on Computer Vision (ACCV)}, 2026.

\bibitem[Deng and Zheng(2021)]{deng_CVPR_2021}
Weijian Deng and Liang Zheng.
\newblock Are labels always necessary for classifier accuracy evaluation?
\newblock In \emph{IEEE/CVF Conference on Computer Vision and Pattern
  Recognition (CVPR)}, pages 15069--15078, 2021.

\bibitem[Du et~al.(2022)Du, Wang, Cai, and Li]{du_ICLR_2022}
Xuefeng Du, Zhaoning Wang, Mu Cai, and Yixuan Li.
\newblock {VOS}: Learning what you don't know by virtual outlier synthesis.
\newblock In \emph{International Conference on Learning Representations
  (ICLR)}, 2022.

\bibitem[Garg et~al.(2022)Garg, Balakrishnan, Lipton, Neyshabur, and
  Sedghi]{garg_ICLR_2022}
Saurabh Garg, Sivaraman Balakrishnan, Zachary~C. Lipton, Behnam Neyshabur, and
  Hanie Sedghi.
\newblock Leveraging unlabeled data to predict out-of-distribution performance.
\newblock In \emph{International Conference on Learning Representations
  (ICLR)}, 2022.

\bibitem[Hall et~al.(2020)Hall, Dayoub, Skinner, Zhang, Miller, Corke,
  Carneiro, Angelova, and S{\"u}nderhauf]{hall_WACV_2020}
David Hall, Feras Dayoub, John Skinner, Haoyang Zhang, Dimity Miller, Peter
  Corke, Gustavo Carneiro, Anelia Angelova, and Niko S{\"u}nderhauf.
\newblock Probabilistic object detection: Definition and evaluation.
\newblock In \emph{IEEE Winter Conference on Applications of Computer Vision
  (WACV)}, pages 1020--1029, 2020.

\bibitem[Hendrycks and Dietterich(2019)]{hendrycks_ICLR_2019}
Dan Hendrycks and Thomas Dietterich.
\newblock Benchmarking neural network robustness to common corruptions and
  perturbations.
\newblock In \emph{International Conference on Learning Representations
  (ICLR)}, 2019.

\bibitem[Hendrycks and Gimpel(2017)]{hendrycks_ICLR_2017}
Dan Hendrycks and Kevin Gimpel.
\newblock A baseline for detecting misclassified and out-of-distribution
  examples in neural networks.
\newblock In \emph{International Conference on Learning Representations
  (ICLR)}, 2017.

\bibitem[K{\"u}ppers et~al.(2020)K{\"u}ppers, Kronenberger, Shantia, and
  Haselhoff]{kuppers_CVPRW_2020}
Fabian K{\"u}ppers, Jan Kronenberger, Amirhossein Shantia, and Anselm
  Haselhoff.
\newblock Multivariate confidence calibration for object detection.
\newblock In \emph{IEEE/CVF Conference on Computer Vision and Pattern
  Recognition (CVPR) Workshops}, pages 1322--1330, 2020.

\bibitem[Lee et~al.(2023)Lee, Varghese, and Vandierendonck]{lee_WACV_2023}
JunKyu Lee, Blesson Varghese, and Hans Vandierendonck.
\newblock {ROMA}: Run-time object detection to maximize real-time accuracy.
\newblock In \emph{IEEE/CVF Winter Conference on Applications of Computer
  Vision (WACV)}, pages 6394--6403, 2023.

\bibitem[Liang et~al.(2018)Liang, Li, and Srikant]{liang_ICLR_2018}
Shiyu Liang, Yixuan Li, and R. Srikant.
\newblock Enhancing the reliability of out-of-distribution image detection in
  neural networks.
\newblock In \emph{International Conference on Learning Representations
  (ICLR)}, 2018.

\bibitem[Lin et~al.(2014)Lin, Maire, Belongie, Hays, Perona, Ramanan,
  Doll{\'a}r, and Zitnick]{lin_ECCV_2014}
Tsung-Yi Lin, Michael Maire, Serge Belongie, James Hays, Pietro Perona, Deva
  Ramanan, Piotr Doll{\'a}r, and C.~Lawrence Zitnick.
\newblock Microsoft {COCO}: Common objects in context.
\newblock In \emph{European Conference on Computer Vision (ECCV)}, pages
  740--755, 2014.

\bibitem[Liu et~al.(2020)Liu, Wang, Owens, and Li]{liu_NeurIPS_2020}
Weitang Liu, Xiaoyun Wang, John Owens, and Yixuan Li.
\newblock Energy-based out-of-distribution detection.
\newblock In \emph{Advances in Neural Information Processing Systems
  (NeurIPS)}, pages 21464--21475, 2020.

\bibitem[Michaelis et~al.(2019)Michaelis, Mitzkus, Geirhos, Rusak, Bringmann,
  Ecker, Bethge, and Brendel]{michaelis_NeurIPSW_2019}
Claudio Michaelis, Benjamin Mitzkus, Robert Geirhos, Evgenia Rusak, Oliver
  Bringmann, Alexander~S. Ecker, Matthias Bethge, and Wieland Brendel.
\newblock Benchmarking robustness in object detection: Autonomous driving when
  winter is coming.
\newblock In \emph{NeurIPS Workshop on Machine Learning for Autonomous
  Driving}, 2019.

\bibitem[Miller et~al.(2019)Miller, S{\"u}nderhauf, Zhang, Hall, and
  Dayoub]{miller_CVPRW_2019}
Dimity Miller, Niko S{\"u}nderhauf, Haoyang Zhang, David Hall, and Feras
  Dayoub.
\newblock Benchmarking sampling-based probabilistic object detectors.
\newblock In \emph{IEEE/CVF Conference on Computer Vision and Pattern
  Recognition (CVPR) Workshops}, pages 42--45, 2019.

\bibitem[Mittal et~al.(2012)Mittal, Moorthy, and Bovik]{mittal_TIP_2012}
Anish Mittal, Anush~Krishna Moorthy, and Alan~Conrad Bovik.
\newblock No-reference image quality assessment in the spatial domain.
\newblock \emph{IEEE Transactions on Image Processing}, 21\penalty0
  (12):\penalty0 4695--4708, 2012.

\bibitem[Mittal et~al.(2013)Mittal, Soundararajan, and Bovik]{mittal_SPL_2013}
Anish Mittal, Rajiv Soundararajan, and Alan~C. Bovik.
\newblock Making a ``completely blind'' image quality analyzer.
\newblock \emph{IEEE Signal Processing Letters}, 20\penalty0 (3):\penalty0
  209--212, 2013.

\bibitem[Oksuz et~al.(2023)Oksuz, Joy, and Dokania]{oksuz_CVPR_2023}
Kemal Oksuz, Tom Joy, and Puneet~K. Dokania.
\newblock Towards building self-aware object detectors via reliable uncertainty
  quantification and calibration.
\newblock In \emph{IEEE/CVF Conference on Computer Vision and Pattern
  Recognition (CVPR)}, pages 9263--9274, 2023.

\bibitem[Pathiraja et~al.(2023)Pathiraja, Gunawardhana, and
  Khan]{pathiraja_CVPR_2023}
Bimsara Pathiraja, Malitha Gunawardhana, and Muhammad~Haris Khan.
\newblock Multiclass confidence and localization calibration for object
  detection.
\newblock In \emph{IEEE/CVF Conference on Computer Vision and Pattern
  Recognition (CVPR)}, pages 19734--19743, 2023.

\bibitem[Shekar et~al.(2021)Shekar, Gou, Ren, and Wendt]{shekar_IJCV_2021}
Arvind~Kumar Shekar, Liang Gou, Liu Ren, and Axel Wendt.
\newblock Label-free robustness estimation of object detection {CNN}s for
  autonomous driving applications.
\newblock \emph{International Journal of Computer Vision}, 129\penalty0
  (4):\penalty0 1185--1201, 2021.

\bibitem[Wang et~al.(2024)Wang, Chen, Liu, Chen, Lin, Han, and
  Ding]{wang_NeurIPS_2024}
Ao Wang, Hui Chen, Lihao Liu, Kai Chen, Zijia Lin, Jungong Han, and Guiguang
  Ding.
\newblock {YOLOv10}: Real-time end-to-end object detection.
\newblock In \emph{Advances in Neural Information Processing Systems
  (NeurIPS)}, 2024.

\bibitem[Yoo et~al.(2025)Yoo, Kwon, Hwang, and Lee]{yoo_ICCV_2025}
Seungju Yoo, Hyuk Kwon, Joong-Won Hwang, and Kibok Lee.
\newblock Automated model evaluation for object detection via prediction
  consistency and reliability.
\newblock In \emph{IEEE/CVF International Conference on Computer Vision
  (ICCV)}, pages 19764--19773, 2025.

\bibitem[Yu et~al.(2020)Yu, Chen, Wang, Xian, Chen, Liu, Madhavan, and
  Darrell]{yu_CVPR_2020}
Fisher Yu, Haofeng Chen, Xin Wang, Wenqi Xian, Yingying Chen, Fangchen Liu,
  Vashisht Madhavan, and Trevor Darrell.
\newblock {BDD100K}: A diverse driving dataset for heterogeneous multitask
  learning.
\newblock In \emph{IEEE/CVF Conference on Computer Vision and Pattern
  Recognition (CVPR)}, pages 2636--2645, 2020.

\end{thebibliography}
}

\section*{LLM Usage Statement}
The research question, the concept of detector operating behavior inference, and the overall approach were developed by the authors. Large language models were used in the subsequent work: LLM-based coding agents implemented and ran follow-up experiments and evaluations according to specifications and protocols set and reviewed by the authors, and an LLM assistant supported the analysis of results, the coordination and refinement of experiments, and the editing of the manuscript. The authors checked the results and the text and take full responsibility for all content.

\clearpage
\onecolumn
\begin{center}
{\Large\textbf{\thetitle}}\\[0.5em]
{\large Supplementary Material}
\end{center}
\vspace{0.5em}
\setcounter{section}{0}\setcounter{figure}{0}\setcounter{table}{0}\setcounter{equation}{0}
\renewcommand{\thesection}{S\arabic{section}}
\renewcommand{\thefigure}{S\arabic{figure}}
\renewcommand{\thetable}{S\arabic{table}}
\renewcommand{\theequation}{S\arabic{equation}}
\renewcommand{\theHsection}{S\arabic{section}}
\renewcommand{\theHfigure}{S\arabic{figure}}
\renewcommand{\theHtable}{S\arabic{table}}
\renewcommand{\theHequation}{S\arabic{equation}}

This supplementary material provides additional implementation details, extended experimental results, and supporting analyses for the main paper. In particular, it includes (i) dataset construction, training sampling, and the canonical proxy-regime nomenclature; (ii) additional analyses of the induced behavior space and the PC1 baseline; (iii) full ROC, FROC, PR, and threshold-domain visualizations; (iv) detailed operational results and a comparison between regime-curve utility and true per-image execution; (v) failure-case and held-out-regime analyses; (vi) knot-placement details; (vii) a detector-generalization study on YOLOv10-m and YOLO11-m, including a low-threshold failure case; (viii) exploratory and pre-specified follow-up studies that compare the functional decoder with regime-supervised lookups and regime-label-free alternatives; (ix) the pre-specified continuous and compositional probing experiments (RQ5); and (x) natural-condition studies on BDD100K, including data preparation.

\section{Additional Experimental Details}
\label{S:sec:details}
\paragraph{Dataset split and expansion.} The monitor training pool is built from the person-focused COCOMini subset: of 8,706 base images, a deterministic 10\% subset (870) is reserved for validation, leaving 7,836 training images. The 1,650-image test pool is drawn from the regular COCO validation side after the same person-focused filtering. Every base image has 52 stored variants (pristine plus severities 3--5 of 17 degradations), giving 407,472 training, 45,240 validation, and 85,800 test variants. The splits share no base image.

\paragraph{Training samples.} Each epoch uses a fixed set of 7,836 variants per regime (141,048 in total), selected once with a fixed seed per regime. Pristine uses all training base images. Each degraded regime uses all three severity variants of 2,612 randomly selected base images. The selection is identical across epochs and seeds; seeds vary the monitor initialization and the data order.

\paragraph{Controlled probing and within-detector similarity.}
The constructed regimes are deliberately used as a controlled identifiability study rather than as a claim about the complete structure of deployment data. For the fixed detector, each intervention in data space acts as an experimental probe: evaluating the detector on that probe reveals one sample of its effective functional state space. The present scheme is structured rather than random because named transformations and severities provide repeatable coverage while preserving semantic annotations. The same formulation could sample conditions stochastically; only the probing mechanism would change. Proxy-regime identities organize the probes and attach distribution-level targets, but the monitor does not predict them.

The state-space interpretation is functional, not parametric: the detector weights remain fixed. The possibility of inducing operating variation through either data or parameter changes motivated our choice of data-space probes, but no correspondence between those intervention spaces is assumed or tested. Within the fixed detector, two sampled data conditions are approximately behaviorally similar when their threshold-dependent responses are close under a chosen metric and tolerance, even if their visual or semantic identities differ. The strong PC1 result should be read in this spirit. It is a \emph{representation-dependent} diagnostic of the deliberately sampled behavior space: the original behavior vector emphasizes mutually related recall and quality quantities. Balancing recall and false-positive behavior reveals additional structure. This neither weakens the inverse-inference proof of concept nor establishes a universal low-dimensional geometry for natural deployment shifts.

\paragraph{Discrete condition baselines and functional behavior.}
A finite condition vocabulary provides a useful diagnostic discretization of the sampled functional state space: if the constructed regime identity is known, one can retrieve a validation-calibrated operating point or a stored behavior template. DOBI uses the same regime structure only to organize controlled probes and supervision, while learning the inverse map from local detector evidence to the threshold-indexed functional state itself. The Regime-ID baseline therefore quantifies how much operating-point information is available from perfect condition identity in this controlled setting. Its strong result confirms that the present signal is predominantly regime-level, while the curve-reconstruction results show that DOBI exposes a richer object than a single condition-specific operating point.

\paragraph{Regime nomenclature.}
To avoid ambiguity between internal identifiers, figure labels, and prose, \Cref{S:tab:regime_names} defines the canonical display names used throughout the paper and supplement. All degraded monitor regimes pool severities 3--5; the pristine regime contains the undegraded reference images.

\begin{table}[H]
\centering
\caption{Canonical proxy-regime names. Internal identifiers are shown for reproducibility; the display names in the middle column are used consistently in text, figures, and tables.}
\label{S:tab:regime_names}
\small
\begin{tabular}{lll}
\toprule
Internal ID & Display name & Family \\
\midrule
\texttt{pristine} & Pristine & Reference \\
\texttt{brightness\_3\_5} & Brightness & Photometric \\
\texttt{contrast\_3\_5} & Contrast & Photometric \\
\texttt{defocus\_blur\_3\_5} & Defocus blur & Blur \\
\texttt{fog\_3\_5} & Fog & Weather \\
\texttt{frost\_3\_5} & Frost & Weather \\
\texttt{gaussian\_blur\_3\_5} & Gaussian blur & Blur \\
\texttt{gaussian\_noise\_3\_5} & Gaussian noise & Noise \\
\texttt{glass\_blur\_3\_5} & Glass blur & Blur \\
\texttt{impulse\_noise\_3\_5} & Impulse noise & Noise \\
\texttt{jpeg\_compression\_3\_5} & JPEG & Digital \\
\texttt{motion\_blur\_3\_5} & Motion blur & Blur \\
\texttt{pixelate\_3\_5} & Pixelate & Digital \\
\texttt{saturate\_3\_5} & Saturate & Photometric \\
\texttt{shot\_noise\_3\_5} & Shot noise & Noise \\
\texttt{snow\_3\_5} & Snow & Weather \\
\texttt{spatter\_3\_5} & Spatter & Weather \\
\texttt{speckle\_noise\_3\_5} & Speckle noise & Noise \\
\bottomrule
\end{tabular}
\end{table}

\paragraph{Area errors.} \Cref{S:tab:area} complements the MAD results of the main paper.
\begin{table}[H]
\centering
\caption{Absolute area errors (macro over 18 test regimes; learned models: mean over three seeds).}
\label{S:tab:area}
\begin{tabular}{lccc}
\toprule
Model & $|\Delta$AUROC$|$ & $|\Delta$AFROC$|$ & $|\Delta$AUPRC$|$ \\
\midrule
Mean behavior & 0.1079 & 0.2569 & 0.1266 \\
Linear & 0.0407 & 0.0665 & 0.0524 \\
Adaptive PCHIP & 0.0365 & 0.0673 & 0.0481 \\
\bottomrule
\end{tabular}

\end{table}

\paragraph{Behavior-space parameterizations.} The behavior-space PCA of the main paper uses the regime vector $[\mathrm{TPR}(\tau),\mathrm{Prec}(\tau),F_1(\tau),\log(1+\mathrm{FPPI}(\tau))/c]$ with fixed FPPI normalization; PC1 explains 96.6\% of the training-regime variance. The PC1 monitor baseline instead uses $[\mathrm{TPR},\mathrm{FPR},\log(1+\mathrm{FPPI}),\mathrm{Prec}]$ (\Cref{S:tab:pc1}), for which PC1 explains 88.0\%. These values arise from different coordinate systems; explained variance changes with redundant components and their scaling. Precision and $F_1$ share underlying TP/FP counts with TPR and FPPI and should not be interpreted as independent behavioral degrees of freedom.

\begin{table}[H]
\centering
\caption{PC1 baseline representation (PCA fitted on the 18 training regimes). Reconstruction RMSE uses the exact PC1 coefficient of each regime (seed independent).}
\label{S:tab:pc1}
\begin{tabular}{ccccc}
\toprule
PC1 & PC2 & PC3 & Train RMSE & Test RMSE \\
\midrule
88.0\% & 10.8\% & 0.7\% & 0.0294 & 0.0297 \\
\bottomrule
\end{tabular}

\end{table}

The PC1 monitor baseline predicts one coefficient from local features, while the diagnostic \emph{exact projection} uses the test target to determine that coefficient and is therefore not deployable. For the PC1-monitor vectorization, PC1 explains 88.0\% of the training-regime variance (PC2 adds 10.8\%); exact one-dimensional reconstruction has train/test RMSE 0.0294/0.0297. The baseline is an informative complexity control, not evidence that all possible detector behaviors are one-dimensional.

\paragraph{Sensitivity to metric weighting.}
A further analysis of the existing, labeled behavior targets separates the correlated output statistics (\Cref{S:tab:pc_weighting}). TPR alone varies predominantly along one axis (98.5\% PC1), as does FPPI alone (91.7\% PC1). However, the corresponding regime-level coefficients correlate only 0.52: degradation-induced recall and false-positive responses need not move together. Giving TPR and FPPI equal weight yields 69\% on PC1 and 29\% on PC2, rather than the 96.6\% reported for the main-paper representation. The latter also includes precision and $F_1$, both derived from TP/FP counts, and consequently emphasizes overlapping quality information. These PCA values refer to the 18 training regimes and their specified feature scaling; they are not estimates of the dimensionality of all deployment conditions.

\begin{table}[H]
\centering
\caption{Representation dependence of functional PCA on the existing training-regime targets. Each row is a separate PCA fit and should only be interpreted in its own normalized coordinate system. Values in the last two columns are percentages of variance. The original figure vector and the PC1-monitor vector differ.}
\label{S:tab:pc_weighting}
\small
\begin{tabular}{lrr}
\toprule
Behavior representation & PC1 & PC2 \\
\midrule
Original main-paper vector & 96.6 & 2.4 \\
PC1-monitor vector & 88.0 & 10.8 \\
TPR only & 98.5 & -- \\
FPPI only & 91.7 & -- \\
TPR + FPPI (balanced) & 69 & 29 \\
\bottomrule
\end{tabular}
\end{table}

\paragraph{Interpretation and operational consequence.}
The dominant direction tracks overall detection-quality loss (reported correlation with regime-optimal $F_1$: $-0.998$), whereas another direction captures false-positive behavior, especially at low thresholds (reported correlation $0.91$). Pixelate, Motion blur, Snow, and several noise conditions deviate most strongly from a scalar approximation. A diagnostic projection of the \emph{true test curves} onto the learned one-PC basis achieves macro $F_1$ regret about 0.0014, compared with 0.0083 for the image-independent mean-behavior policy. This is an oracle-coefficient reconstruction test, not a deployable monitor result or a guarantee for all utilities.

The models also trade reconstruction fidelity across curve types: PCHIP has lower ROC and FROC MAD, whereas PC1 has lower PR MAD in the reported comparison. Under true per-image $F_1$ execution, PCHIP has a macro advantage of 0.0003 (not statistically established) and a small micro advantage of about 0.0007. Yet a perfect one-PC projection can lose critical FPPI information: the reported diagnostic FPPI-policy analysis observes budget exceedances in 13--15 of 18 regimes; including PC2 reduces this to 9--10, without establishing a calibrated safety guarantee. As further discussed in \Cref{S:sec:supp_fppi}, the full functional prediction is not uniformly superior to PC1, but it provides information beyond a single quality coordinate for some operating policies.

\section{Full Functional Behavior}
\label{S:sec:full}
\begin{figure}[H]
\centering
\includegraphics[width=\textwidth]{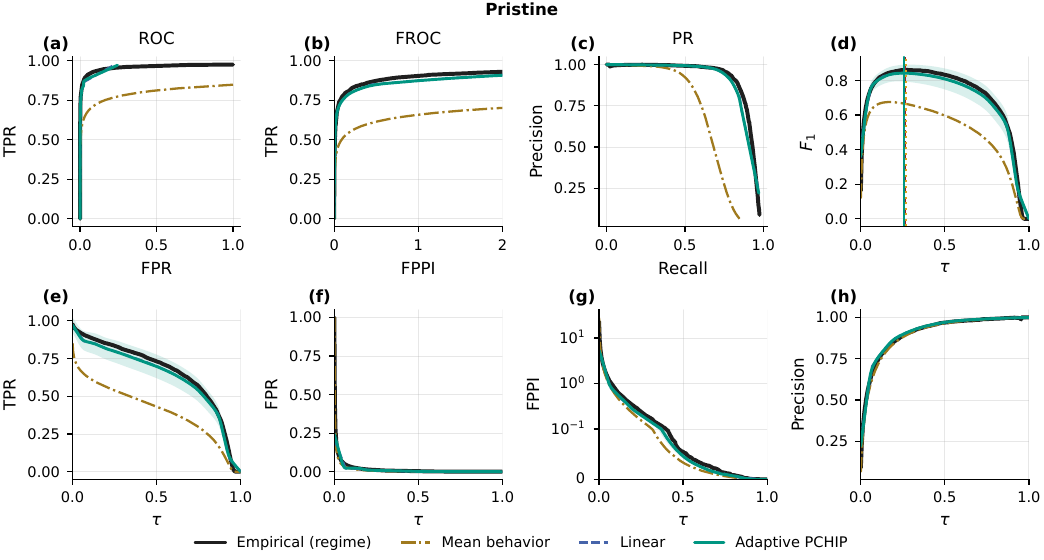}
\caption{Full functional behavior, Pristine. (a)~ROC, (b)~FROC, (c)~PR, (d)~$F_1(\tau)$ with fixed pristine, mean adaptive PCHIP, and regime-oracle thresholds, (e--h)~TPR, FPR, FPPI (symlog), and precision versus $\tau$. ROC and FROC are shown over the empirically realized operating segment induced by the finite set of stored detector candidates and the evaluated threshold range; no artificial endpoints are appended.}
\label{S:fig:full_pristine}
\end{figure}

\begin{figure}[H]
\centering
\includegraphics[width=\textwidth]{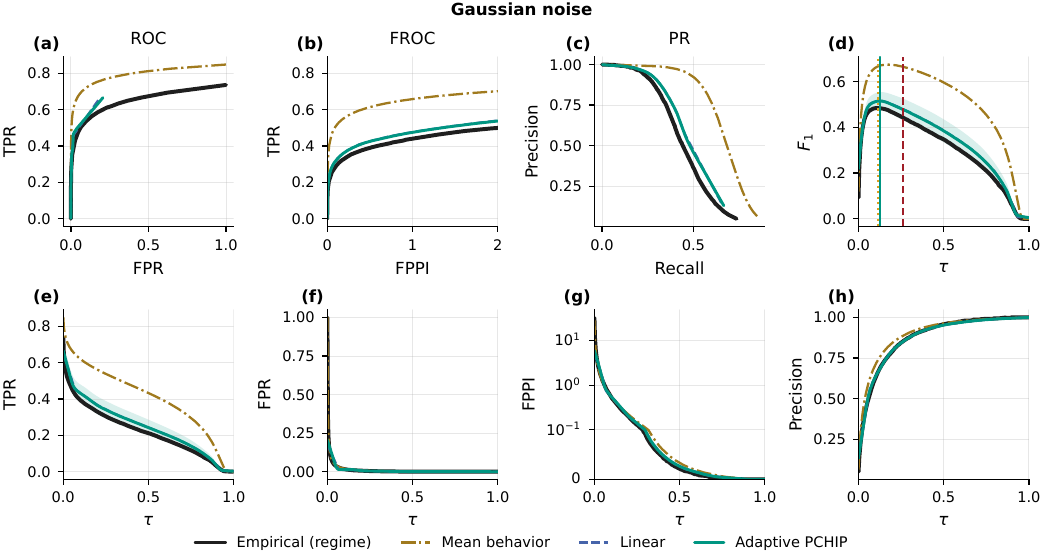}
\caption{Full functional behavior, Gaussian noise (severity 3--5). (a)~ROC, (b)~FROC, (c)~PR, (d)~$F_1(\tau)$ with fixed pristine, mean adaptive PCHIP, and regime-oracle thresholds, (e--h)~TPR, FPR, FPPI (symlog), and precision versus $\tau$. ROC and FROC are shown over the empirically realized operating segment induced by the finite set of stored detector candidates and the evaluated threshold range; no artificial endpoints are appended.}
\label{S:fig:full_noise}
\end{figure}

\begin{figure}[H]
\centering
\includegraphics[width=\textwidth]{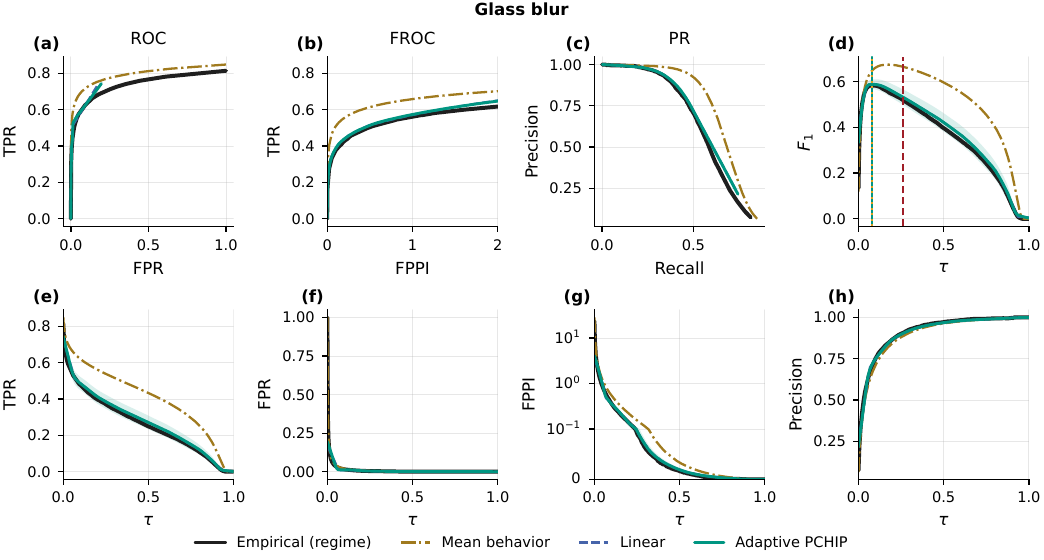}
\caption{Full functional behavior, Glass blur (severity 3--5). (a)~ROC, (b)~FROC, (c)~PR, (d)~$F_1(\tau)$ with fixed pristine, mean adaptive PCHIP, and regime-oracle thresholds, (e--h)~TPR, FPR, FPPI (symlog), and precision versus $\tau$. ROC and FROC are shown over the empirically realized operating segment induced by the finite set of stored detector candidates and the evaluated threshold range; no artificial endpoints are appended.}
\label{S:fig:full_glass}
\end{figure}

\begin{figure}[H]
\centering
\includegraphics[width=\textwidth]{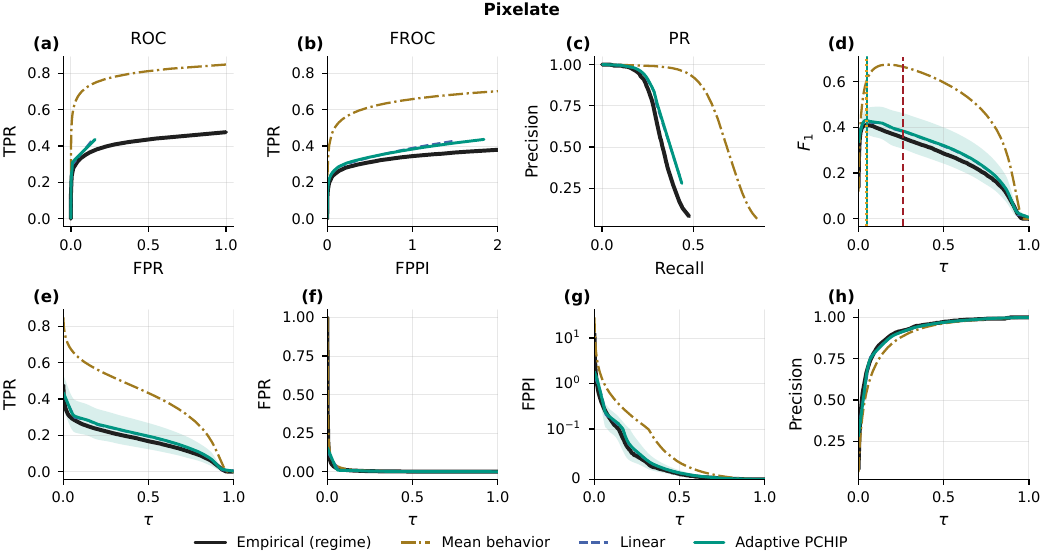}
\caption{Full functional behavior, Pixelate (severity 3--5). (a)~ROC, (b)~FROC, (c)~PR, (d)~$F_1(\tau)$ with fixed pristine, mean adaptive PCHIP, and regime-oracle thresholds, (e--h)~TPR, FPR, FPPI (symlog), and precision versus $\tau$. ROC and FROC are shown over the empirically realized operating segment induced by the finite set of stored detector candidates and the evaluated threshold range; no artificial endpoints are appended.}
\label{S:fig:full_pixelate}
\end{figure}

\begin{figure}[H]
\centering
\includegraphics[width=\textwidth]{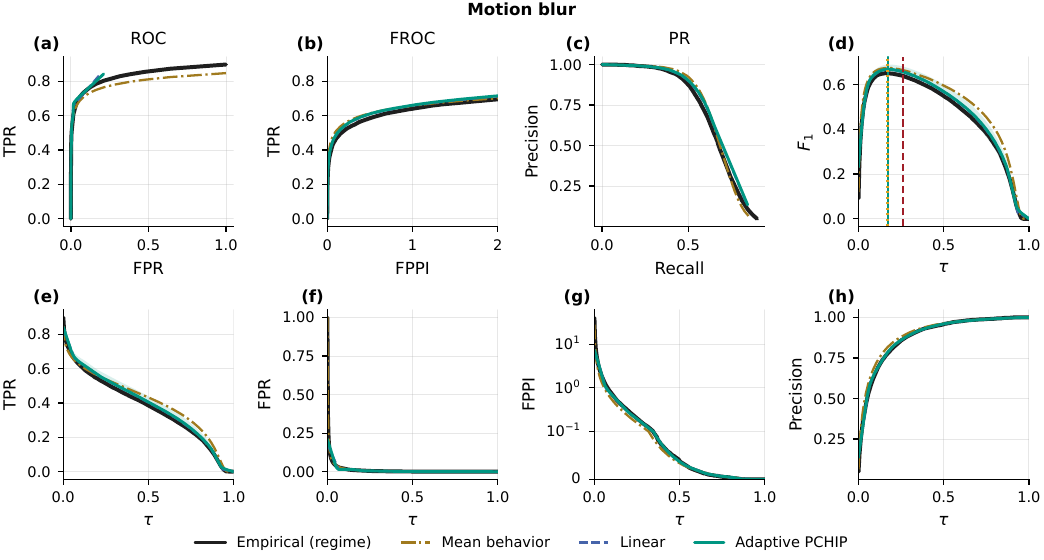}
\caption{Full functional behavior, Motion blur (severity 3--5). (a)~ROC, (b)~FROC, (c)~PR, (d)~$F_1(\tau)$ with fixed pristine, mean adaptive PCHIP, and regime-oracle thresholds, (e--h)~TPR, FPR, FPPI (symlog), and precision versus $\tau$. ROC and FROC are shown over the empirically realized operating segment induced by the finite set of stored detector candidates and the evaluated threshold range; no artificial endpoints are appended.}
\label{S:fig:full_motion}
\end{figure}

\begin{figure}[H]
\centering
\includegraphics[width=\textwidth]{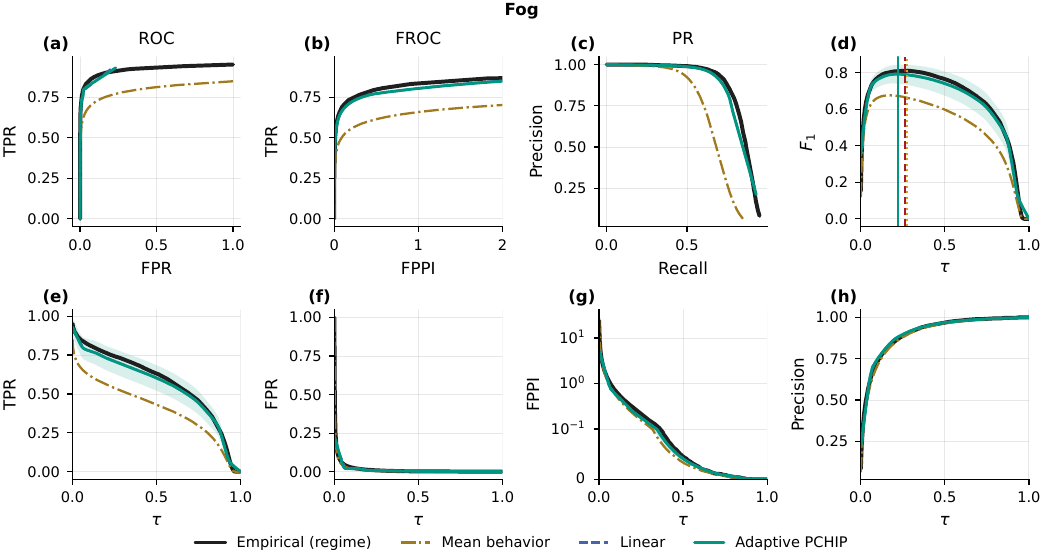}
\caption{Full functional behavior, Fog (severity 3--5). (a)~ROC, (b)~FROC, (c)~PR, (d)~$F_1(\tau)$ with fixed pristine, mean adaptive PCHIP, and regime-oracle thresholds, (e--h)~TPR, FPR, FPPI (symlog), and precision versus $\tau$. ROC and FROC are shown over the empirically realized operating segment induced by the finite set of stored detector candidates and the evaluated threshold range; no artificial endpoints are appended.}
\label{S:fig:full_fog}
\end{figure}

\Cref{S:fig:full_pristine,S:fig:full_noise,S:fig:full_glass,S:fig:full_pixelate,S:fig:full_motion,S:fig:full_fog} show ROC, FROC, PR, and the threshold-domain functions for six representative test regimes. Predictions are averages of the per-image predicted threshold functions over all test images of the regime and over three seeds. Shaded bands (threshold panels) denote one standard deviation across images, centered on the plotted mean. ROC and FROC are shown over the empirically realized operating segment induced by the finite set of stored detector candidates and the evaluated threshold range; no artificial endpoints are appended. FROC is shown on the evaluation domain $\mathrm{FPPI}\in[0,2]$.

\clearpage

\section{Operational Results in Detail}
\label{S:sec:operational}
\begin{figure}[H]
\centering
\includegraphics[width=\textwidth]{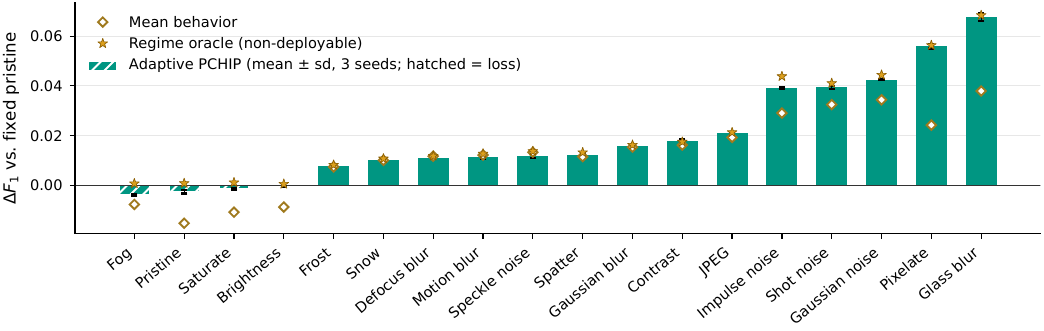}
\caption{Per-regime $F_1$ gain over the fixed pristine threshold under true execution. Bars: adaptive PCHIP; open diamonds: mean-behavior prior; stars: regime oracle. Negative cases are retained.}
\label{S:fig:gain}
\end{figure}

\begin{figure}[H]
\centering
\includegraphics[width=\textwidth]{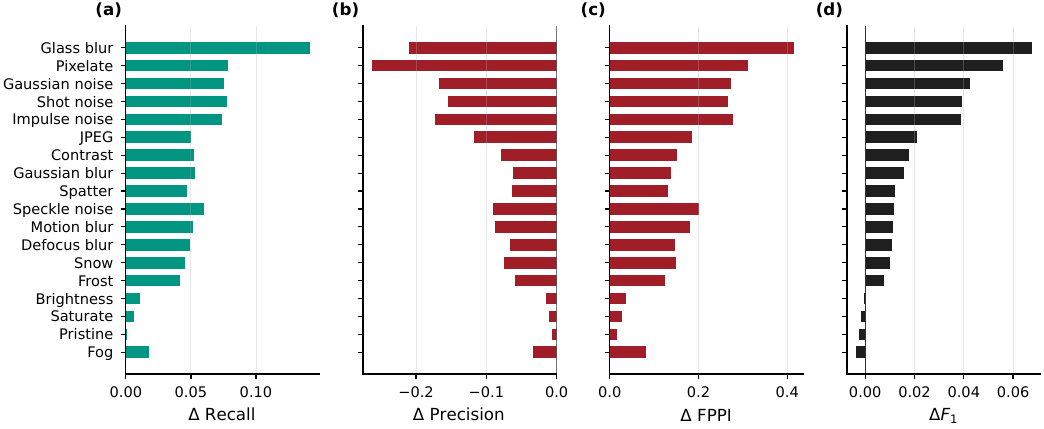}
\caption{Adaptive PCHIP minus fixed pristine per regime (counts summed over all variants, mean over three seeds), sorted by $\Delta F_1$: (a)~recall, (b)~precision, (c)~FPPI, (d)~$F_1$. Every regime gains recall and loses precision; no regime gains by reducing false positives.}
\label{S:fig:tradeoff}
\end{figure}

\begin{figure}[H]
\centering
\includegraphics[width=\textwidth]{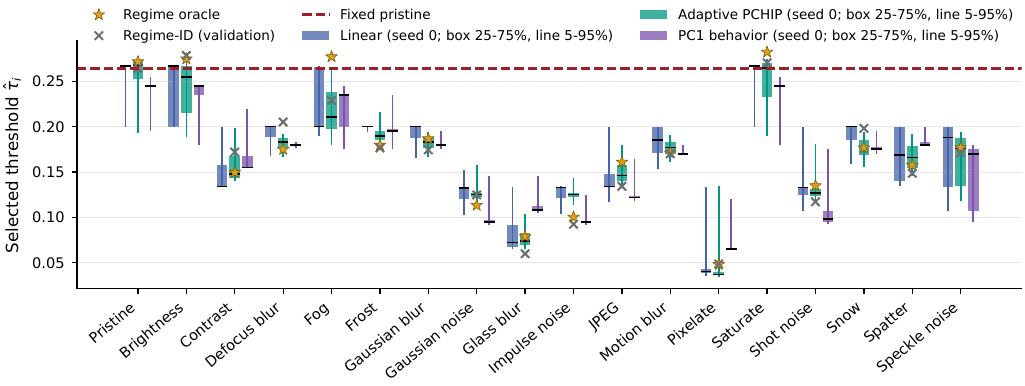}
\caption{Distribution of per-image selected thresholds per regime (seed 0): boxes 25--75\%, lines 5--95\%, black ticks median. Stars: regime oracle; crosses: Regime-ID validation threshold; dashed line: fixed pristine threshold. With linear interpolation many thresholds sit on knots (multiples of 1/15).}
\label{S:fig:taudist}
\end{figure}

\begin{figure}[H]
\centering
\includegraphics[width=0.66\textwidth]{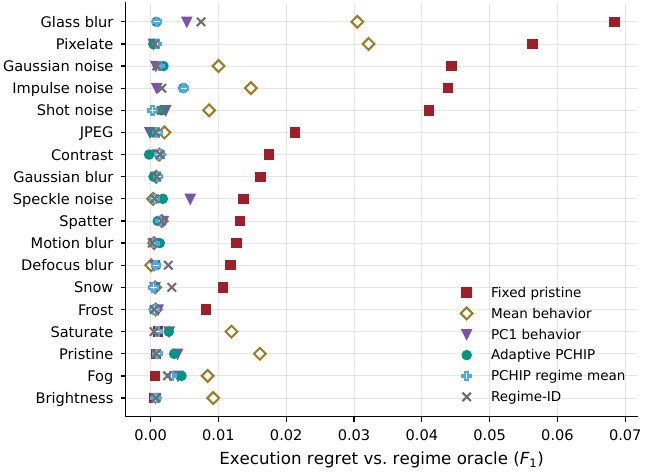}
\caption{Per-regime execution regret with respect to the regime oracle (learned policies: mean over three seeds).}
\label{S:fig:regret}
\end{figure}

Unless stated otherwise, operational results in this section use true per-image threshold execution: each selected threshold is applied only to the detections of its corresponding image before TP/FP/FN counts are aggregated. \Cref{S:tab:opfull} extends the main operational table, while \Cref{S:tab:perregime} lists every regime. \Cref{S:fig:tradeoff} shows how the $F_1$ change arises; adaptation increases recall at the cost of additional false positives in every regime. \Cref{S:fig:taudist} shows the per-image thresholds, and \Cref{S:fig:regret} the per-regime execution regret. Regime-curve utility is reported separately in \Cref{S:sec:modes}. \Cref{S:fig:gain} shows the per-regime $F_1$ gain, and \Cref{S:tab:fppi_policy_supp} the FPPI-budget policies for $B=0.1$; \Cref{S:tab:medium_fppi} extends them to further budgets.

\begin{table}[H]
\centering
\caption{True per-image execution for an FPPI budget $B=0.1$ . $^{\dagger}$Number of test regimes above the budget (model: three-seed mean). $^{\ddagger}$Known regime identity; not deployable. Validation calibration is shared across regimes.}
\label{S:tab:fppi_policy_supp}
\small
\begin{tabular}{lcc}
\toprule
Policy ($B=0.1$) & Macro recall & Violations$^{\dagger}$ \\
\midrule
Fixed (val., worst case) & 0.438 & 0/18 \\
PCHIP (uncalibrated) & 0.500 & 10.3/18 \\
PCHIP (val.-calibrated) & \textbf{0.462} & \textbf{0/18} \\
Regime-ID (val.)$^{\ddagger}$ & 0.486 & 0/18 \\
\bottomrule
\end{tabular}

\end{table}

\begin{table}[H]
\centering
\caption{True per-image execution on all 85,800 test variants (learned: mean over three seeds). P, R, FPPI and micro $F_1$ are pooled; macro $F_1$, gain and regret average the 18 regimes; $+/-$ counts improved/degraded regimes. $^\dagger$Non-deployable. \emph{PCHIP regime mean} applies the mean PCHIP threshold of each regime and uses the regime grouping.}
\label{S:tab:opfull}
\resizebox{\textwidth}{!}{%
\begin{tabular}{lcccccccc}
\toprule
Policy & P & R & Micro $F_1$ & FPPI & Macro $F_1$ & Gain & Regret & +/$-$ \\
\midrule
Fixed pristine & 0.888 & 0.512 & 0.6494 & 0.148 & 0.6407 & +0.0000 & 0.0212 & 0/0 \\
Mean behavior & 0.817 & 0.556 & 0.6617 & 0.285 & 0.6536 & +0.0129 & 0.0083 & 14/4 \\
Linear & 0.801 & 0.564 & 0.6619 & 0.320 & 0.6600 & +0.0193 & 0.0019 & 14/4 \\
PC1 behavior & 0.793 & 0.567 & 0.6615 & 0.339 & 0.6601 & +0.0194 & 0.0018 & 14/4 \\
Adaptive PCHIP & 0.799 & 0.566 & 0.6622 & 0.327 & 0.6604 & +0.0197 & 0.0016 & 14/4 \\
\midrule
PCHIP regime mean$^\dagger$ & 0.802 & 0.565 & 0.6626 & 0.320 & 0.6607 & +0.0200 & 0.0012 & 15/3 \\
Regime-ID$^\dagger$ & 0.791 & 0.567 & 0.6605 & 0.344 & 0.6604 & +0.0197 & 0.0016 & 15/2 \\
Regime oracle$^\dagger$ & 0.800 & 0.565 & 0.6625 & 0.324 & 0.6620 & +0.0212 & 0.0000 & 18/0 \\
\bottomrule
\end{tabular}
}
\end{table}

\begin{table}[H]
\centering
\caption{Per-regime true execution $F_1$. $\tau^*$: regime-oracle threshold; $\bar{\hat\tau}$: mean adaptive PCHIP threshold; Gain: PCHIP minus fixed pristine.}
\label{S:tab:perregime}
\resizebox{\textwidth}{!}{%
\begin{tabular}{lcccccccc}
\toprule
Regime & $\tau^*$ & $\bar{\hat\tau}$ & Fixed & Mean & PCHIP & Reg.-ID & Oracle & Gain \\
\midrule
Pristine & 0.272 & 0.259 & 0.8619 & 0.8466 & 0.8592 & 0.8619 & 0.8627 & -0.0027 \\
Brightness & 0.275 & 0.245 & 0.8341 & 0.8254 & 0.8337 & 0.8339 & 0.8346 & -0.0004 \\
Contrast & 0.149 & 0.159 & 0.6317 & 0.6479 & 0.6494 & 0.6478 & 0.6492 & +0.0177 \\
Defocus blur & 0.175 & 0.181 & 0.6934 & 0.7051 & 0.7043 & 0.7026 & 0.7052 & +0.0109 \\
Fog & 0.277 & 0.220 & 0.8099 & 0.8021 & 0.8060 & 0.8081 & 0.8105 & -0.0039 \\
Frost & 0.180 & 0.189 & 0.7209 & 0.7283 & 0.7285 & 0.7285 & 0.7290 & +0.0077 \\
Gaussian blur & 0.187 & 0.181 & 0.6915 & 0.7068 & 0.7072 & 0.7068 & 0.7077 & +0.0157 \\
Gaussian noise & 0.113 & 0.128 & 0.4412 & 0.4756 & 0.4837 & 0.4847 & 0.4856 & +0.0425 \\
Glass blur & 0.079 & 0.081 & 0.5176 & 0.5556 & 0.5852 & 0.5786 & 0.5861 & +0.0675 \\
Impulse noise & 0.100 & 0.127 & 0.4471 & 0.4762 & 0.4861 & 0.4893 & 0.4910 & +0.0390 \\
JPEG & 0.161 & 0.148 & 0.5163 & 0.5356 & 0.5372 & 0.5369 & 0.5376 & +0.0209 \\
Motion blur & 0.171 & 0.176 & 0.6398 & 0.6520 & 0.6512 & 0.6523 & 0.6525 & +0.0113 \\
Pixelate & 0.048 & 0.052 & 0.3540 & 0.3782 & 0.4100 & 0.4098 & 0.4104 & +0.0559 \\
Saturate & 0.282 & 0.253 & 0.8486 & 0.8377 & 0.8470 & 0.8492 & 0.8497 & -0.0016 \\
Shot noise & 0.135 & 0.134 & 0.4836 & 0.5160 & 0.5230 & 0.5231 & 0.5246 & +0.0395 \\
Snow & 0.177 & 0.175 & 0.6639 & 0.6739 & 0.6740 & 0.6714 & 0.6746 & +0.0101 \\
Spatter & 0.157 & 0.170 & 0.7039 & 0.7154 & 0.7160 & 0.7154 & 0.7171 & +0.0121 \\
Speckle noise & 0.177 & 0.162 & 0.6738 & 0.6871 & 0.6856 & 0.6869 & 0.6875 & +0.0119 \\
\bottomrule
\end{tabular}
}
\end{table}

\section{Regime-Curve Utility versus True Execution}
\label{S:sec:modes}
The regime-curve utility scores each predicted threshold on the complete empirical $F_1$ curve of the regime, averaged over 1,650 sampled variants per regime. True execution applies the threshold to the detections of its own image and aggregates counts over all variants. \Cref{S:tab:modes} shows that both modes rank the policies identically and agree on the sign of the gain in every regime.

\begin{table}[H]
\centering
\caption{Macro $F_1$ and regret under regime-curve utility (curve) and true per-image execution (exec.).}
\label{S:tab:modes}
\begin{tabular}{lcccccc}
\toprule
Policy & \multicolumn{2}{c}{Macro $F_1$} & \multicolumn{2}{c}{Regret} & Same sign & max $|\Delta F_1|$ \\
 & curve & exec. & curve & exec. & of gain & per regime \\
\midrule
Fixed pristine & 0.6407 & 0.6407 & 0.0212 & 0.0212 & 18/18 & 0.0000 \\
Mean behavior & 0.6536 & 0.6536 & 0.0083 & 0.0083 & 18/18 & 0.0000 \\
Linear & 0.6594 & 0.6600 & 0.0025 & 0.0019 & 18/18 & 0.0034 \\
Adaptive PCHIP & 0.6598 & 0.6604 & 0.0022 & 0.0016 & 18/18 & 0.0035 \\
PC1 behavior & 0.6596 & 0.6601 & 0.0024 & 0.0018 & 18/18 & 0.0026 \\
\bottomrule
\end{tabular}

\end{table}

\section{Fog Failure Case}
\label{S:sec:fog}
\begin{figure}[H]
\centering
\includegraphics[width=\textwidth]{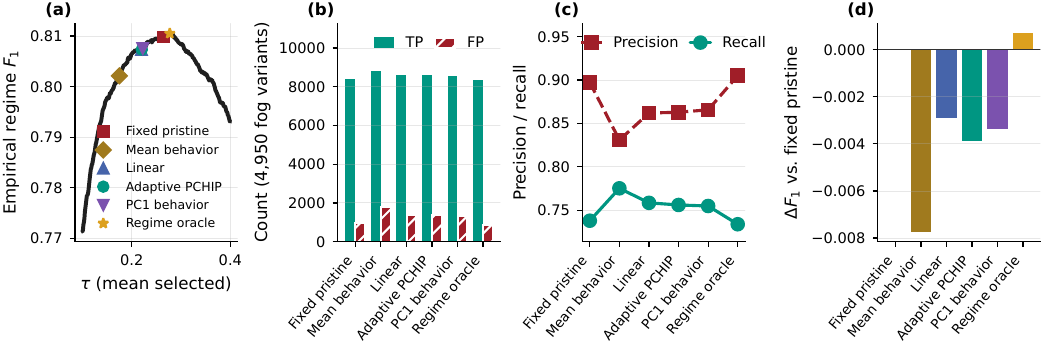}
\caption{Fog under true execution. (a)~Empirical regime $F_1(\tau)$ with the mean threshold selected by each policy. (b)~TP and FP over 4,950 Fog variants. (c)~Precision and recall. (d)~$F_1$ change relative to fixed pristine. The adaptive threshold is too low: the additional recall does not compensate for the added false positives. Linear and PC1 behave similarly, so the failure is not caused by PCHIP interpolation.}
\label{S:fig:fog}
\end{figure}

\section{Held-Out Degradation Types}
\label{S:sec:heldout}
\begin{table}[H]
\centering
\caption{Leave-one-degradation-type-out evaluation (single seed; the held-out type is absent from monitor training and validation). Metrics refer to the held-out regime. $^\ddagger$Regime-curve utility.}
\label{S:tab:heldout}
\resizebox{\textwidth}{!}{%
\begin{tabular}{llccccc}
\toprule
Held-out type & Model & ROC MAD & PR MAD & Gain$^\ddagger$ & Regret$^\ddagger$ & $|\Delta\tau|$ \\
\midrule
Gaussian noise & held-out PCHIP & 0.053 & 0.085 & +0.042 & 0.0026 & 0.019 \\
 & held-out mean behavior & 0.151 & 0.229 & +0.034 & 0.0100 & 0.062 \\
 & in-distribution PCHIP (3 seeds) & 0.047 & 0.062 & +0.043 & 0.0017 & 0.015 \\
Glass blur & held-out PCHIP & 0.094 & 0.161 & +0.032 & 0.0368 & 0.110 \\
 & held-out mean behavior & 0.056 & 0.099 & +0.037 & 0.0312 & 0.097 \\
 & in-distribution PCHIP (3 seeds) & 0.040 & 0.051 & +0.066 & 0.0027 & 0.011 \\
Fog & held-out PCHIP & 0.086 & 0.098 & -0.006 & 0.0066 & 0.078 \\
 & held-out mean behavior & 0.134 & 0.181 & -0.009 & 0.0098 & 0.109 \\
 & in-distribution PCHIP (3 seeds) & 0.049 & 0.054 & -0.003 & 0.0038 & 0.057 \\
\bottomrule
\end{tabular}
}
\end{table}

\section{Knot Placement}
\label{S:sec:knots}
\begin{table}[H]
\centering
\caption{Knot positions of the $K=16$ variants (4 decimals). All-score quantiles concentrate 15 of 16 knots below 0.09, where most of the low-confidence detections lie.}
\label{S:tab:knots}
\small
\begin{tabular}{lp{0.8\linewidth}}
\toprule
Placement & Knots $\tau_1,\ldots,\tau_{16}$ \\
\midrule
uniform & \parbox[t]{0.78\linewidth}{\raggedright\footnotesize 0.0000, 0.0667, 0.1333, 0.2000, 0.2667, 0.3333, 0.4000, 0.4667, 0.5333, 0.6000, 0.6667, 0.7333, 0.8000, 0.8667, 0.9333, 1.0000} \\
all-score quantiles & \parbox[t]{0.78\linewidth}{\raggedright\footnotesize 0.0000, 0.0011, 0.0013, 0.0015, 0.0017, 0.0020, 0.0023, 0.0028, 0.0034, 0.0042, 0.0055, 0.0076, 0.0118, 0.0227, 0.0871, 1.0000} \\
TP-score quantiles & \parbox[t]{0.78\linewidth}{\raggedright\footnotesize 0.0000, 0.0044, 0.0142, 0.0383, 0.0888, 0.1762, 0.2964, 0.4323, 0.5605, 0.6693, 0.7544, 0.8179, 0.8632, 0.8962, 0.9230, 1.0000} \\
\bottomrule
\end{tabular}

\end{table}

\section{Additional Behavior-Inference Evidence}
\label{S:sec:extra_eval}
All analyses in this section use the monitor checkpoints selected by minimum validation loss within the 60-epoch runs (three seeds), the existing detector predictions and features, and the original test variants. No detector or monitor was retrained. The experiments use the same matching protocol, confidence score floor, and threshold-domain targets as the main paper. Full result tables are available from the authors. All new operational results use true per-image threshold execution unless stated otherwise.

\subsection{Aggregation of Local Evidence}
\label{S:sec:supp_batch}
A regime-level behavior function is inferred from individual images. To assess evidence sufficiency, predicted TPR, FPR, FPPI, and precision functions were averaged over batches of $b\in\{1,2,4,8,16,32\}$ independent base images drawn from the same regime (500 draws per size). Multiple variants of the same original image were not treated as independent evidence. The model is unchanged; this is post-hoc averaging of per-image predictions.

\begin{table}[H]
\centering
\caption{PCHIP behavior reconstruction under batch aggregation (macro over 18 regimes, three seeds). ``All'' denotes averaging all images of a regime. Reported threshold errors are distances to the regime oracle, not per-image intrinsic targets.}
\label{S:tab:new_batch}
\begin{tabular}{lrrrr}
\toprule
Batch $b$ & ROC MAD & FROC MAD & PR MAD & $|\hat\tau-\tau^*|$ \\
\midrule
1 & 0.0427 & 0.0340 & 0.0534 & 0.0186 \\
8 & 0.0309 & 0.0256 & 0.0454 & 0.0152 \\
32 & 0.0282 & 0.0242 & 0.0439 & 0.0142 \\
All & 0.0271 & 0.0239 & 0.0435 & 0.0141 \\
\bottomrule
\end{tabular}
\end{table}
\begin{figure}[H]
\centering
\includegraphics[width=\textwidth]{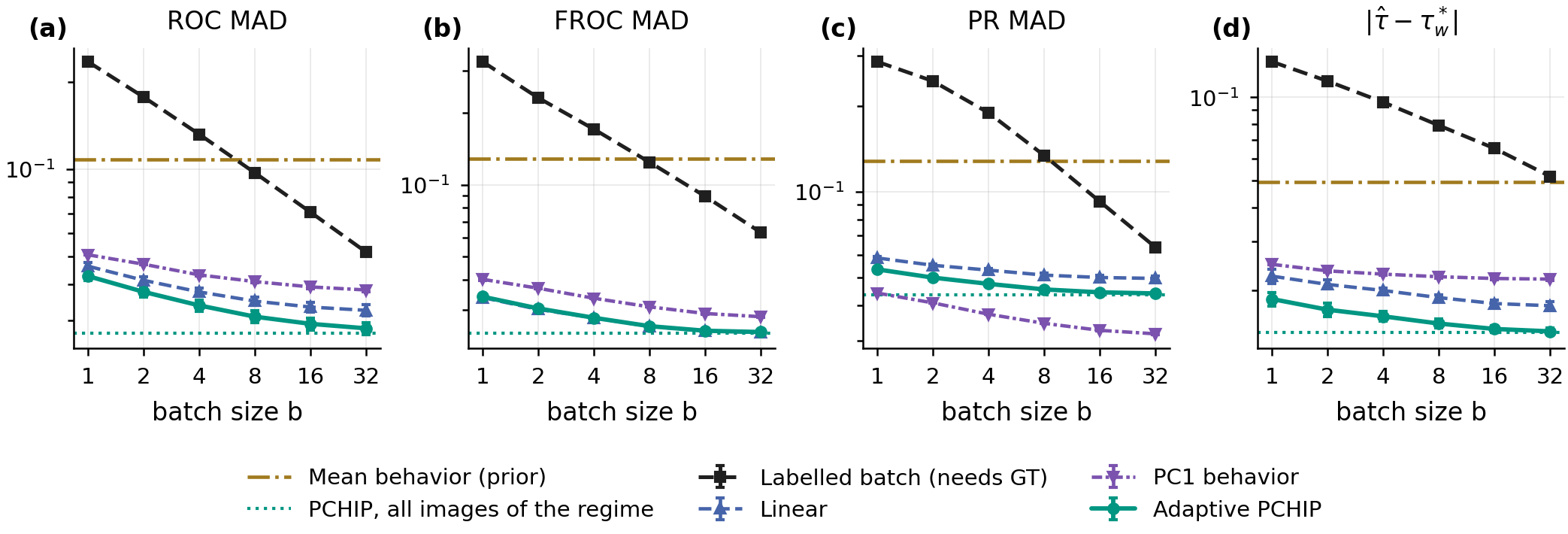}
\caption{Curve error versus number of images used to average the monitor's predictions. Improvement saturates as per-regime bias begins to dominate.}
\label{S:fig:extended_batch}
\end{figure}
Averaging 32 predictions removes most of the reducible sampling error but does not improve strongly shifted cases in which a systematic bias remains. Relative to the mean-behavior prior, one unlabelled monitor observation already reduces substantial prediction error within the trained family. One unlabelled monitor observation is more accurate in macro average than an empirical curve estimated directly from 32 newly labelled images, but this is \emph{not} a comparison at equal supervision cost: the monitor was trained beforehand using many labelled proxy examples. A single image supplies probabilistic evidence of the sampled latent condition; it does not uniquely identify an arbitrary unknown data distribution.

\subsection{Functional Similarity and Distance Geometry}
\label{S:sec:supp_equiv}
For each condition we compare threshold-indexed response vectors with a distance standardized using training data. The tolerance $\varepsilon$ is determined from between-split measurement variation on train/validation conditions; it is not fitted to the test pairs. Functional equivalence is always \emph{approximate}, and the identified pairs depend on the distance metric. Two conditions may differ visually while remaining near each other functionally, without establishing that every threshold-dependent statistic is identical.

\begin{figure}[H]
\centering
\includegraphics[width=\textwidth]{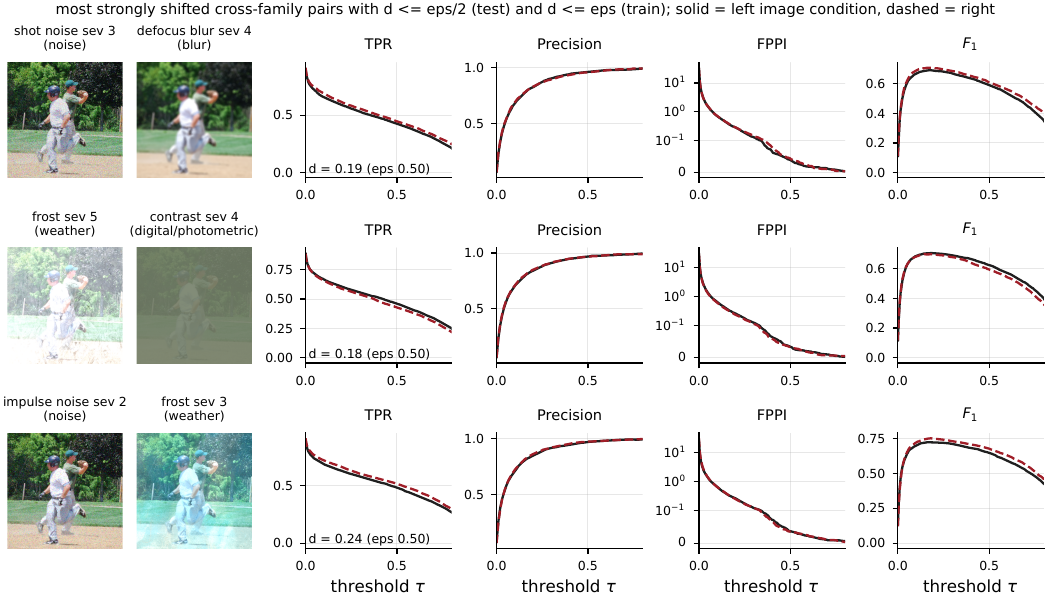}
\caption{Examples of visually distinct cross-family data-space interventions inducing similar empirical detector behavior (threshold-domain TPR, precision, FPPI, and $F_1$). Solid and dashed responses are from the respective conditions. The comparisons are made on disjoint test images using a measurement-based tolerance.}
\label{S:fig:extended_examples}
\end{figure}
\begin{figure}[H]
\centering
\includegraphics[width=0.89\textwidth]{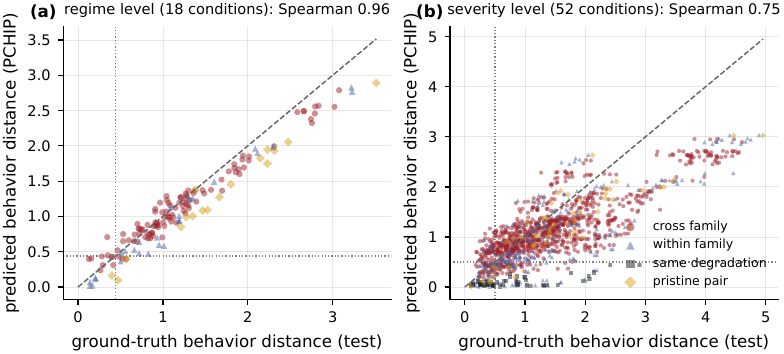}
\caption{Empirical versus predicted behavior distances at regime level (left; Spearman 0.96) and severity level (right; Spearman 0.75). The monitor captures coarse functional structure more accurately than severity-specific differences.}
\label{S:fig:extended_distances}
\end{figure}
Cross-family affiliation alone is weak evidence for functional closeness (AUROC approximately 0.51--0.57, depending on metric). Under the standardized distance, five of 84 shifted cross-family regime pairs are approximately equivalent, compared with 182 of 1,292 severity-level pairs. The result is strongly sensitive to the metric and tolerance: for example, the geometric ROC/FROC/PR metric identifies no equivalent regime-level pairs. These findings indicate approximate detector-specific similarity, not exact equivalence.

\subsection{FPPI-Constrained Operating Policies}
\label{S:sec:supp_fppi}
For a predefined FPPI budget $B$, the monitor selects the threshold maximizing predicted TPR subject to predicted FPPI no greater than an internal budget $\kappa B$. The common safety factor $\kappa$ is selected using validation data separately for each model and budget; calibration is not allowed to use test labels. A fixed worst-case validation policy is calibrated to satisfy the same budget in every validation regime. Regime-ID and test oracle are non-deployable reference policies. Test scores are obtained through true per-image execution.

\begin{table}[H]
\centering
\caption{Recall at different FPPI budgets. Calibrated PCHIP and the fixed worst-case policy had no observed violations in the 18 test regimes. $\Delta R$ is the recall advantage of calibrated PCHIP; paired-bootstrap 95\% intervals were obtained using base-image resampling.}
\label{S:tab:supp_fppi_budgets}
\begin{tabular}{rrrrl}
\toprule
Budget & Fixed recall & PCHIP recall & $\Delta R$ & 95\% CI \\
\midrule
0.05 & 0.371 & 0.398 & +0.027 & [0.026, 0.028] \\
0.10 & 0.438 & 0.462 & +0.023 & [0.022, 0.024] \\
0.20 & 0.513 & 0.531 & +0.017 & [0.017, 0.018] \\
0.50 & 0.581 & 0.593 & +0.012 & [0.011, 0.012] \\
\bottomrule
\end{tabular}
\end{table}
\begin{figure}[H]
\centering
\includegraphics[width=\textwidth]{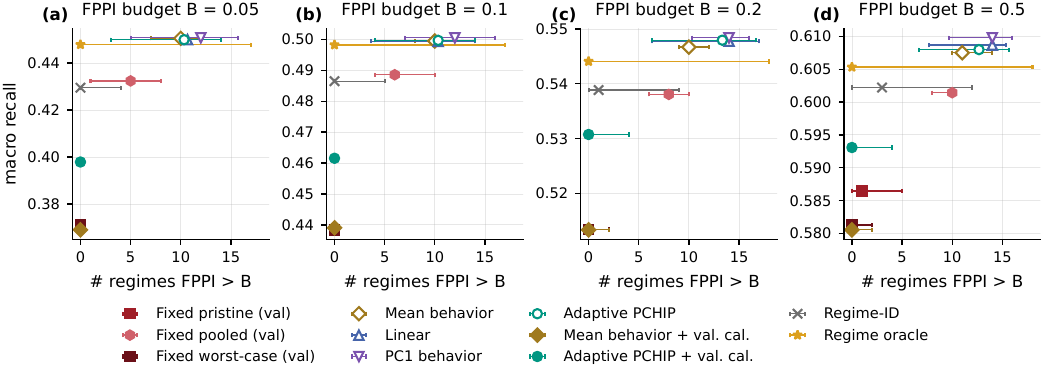}
\caption{Recall--constraint trade-off under predicted versus validation-calibrated FPPI policies. Uncalibrated methods appear competitive in recall but can exceed per-regime budgets; calibrated policies trade some recall for observed constraint compliance.}
\label{S:fig:extended_fppi}
\end{figure}
Uncalibrated PCHIP exceeds the budget in approximately 10--14 of 18 regimes across budgets, even though its predicted FPPI meets the nominal threshold. After validation calibration, no exceedances were observed across the evaluated test regimes. A separate re-analysis of the existing PC1 and PCHIP checkpoints finds no uniform advantage of the richer decoder: PC1 is slightly better at the tight $B=0.05$ and $B=0.1$ budgets, whereas PCHIP achieves about 0.008--0.009 more recall at $B\geq0.2$ under validation calibration. These model comparisons are conditional on the calibration and the sampled regimes; their exact seed-level statistics are not part of the original FPPI-policy table above. The absence of observed budget exceedances for calibrated PCHIP does \emph{not} constitute a distribution-free guarantee: bootstrap estimates of per-regime violation probability can be appreciable, and previously unseen operating states were not covered by this calibration analysis.

\subsection{Consistency of the Functional Outputs}
\label{S:sec:supp_consistency}
The independently predicted behavior components can be checked against identities of empirical detection curves. If $\rho$ denotes the ground-truth person density per image in a regime, then $\mathrm{Prec}(\tau)=\rho\mathrm{TPR}(\tau)/(\rho\mathrm{TPR}(\tau)+\mathrm{FPPI}(\tau))$ where the denominator is nonzero. A second identity links FPR to FPPI under the evaluation convention. Using the training density $\rho=2.189$, the precision-identity discrepancy is 0.0032 for PCHIP, below the approximately 0.0039 mismatch floor from the train/test density difference. After excluding the unresolved score-floor boundary and anchoring at the first positive knot, the spline heads still exhibit FPR--FPPI inconsistency: 0.058 for PCHIP versus 0.001 for PC1. The threshold floor below $1/15$ is inadequately resolved by the uniform 16-knot representation.
\begin{figure}[H]
\centering
\includegraphics[width=\textwidth]{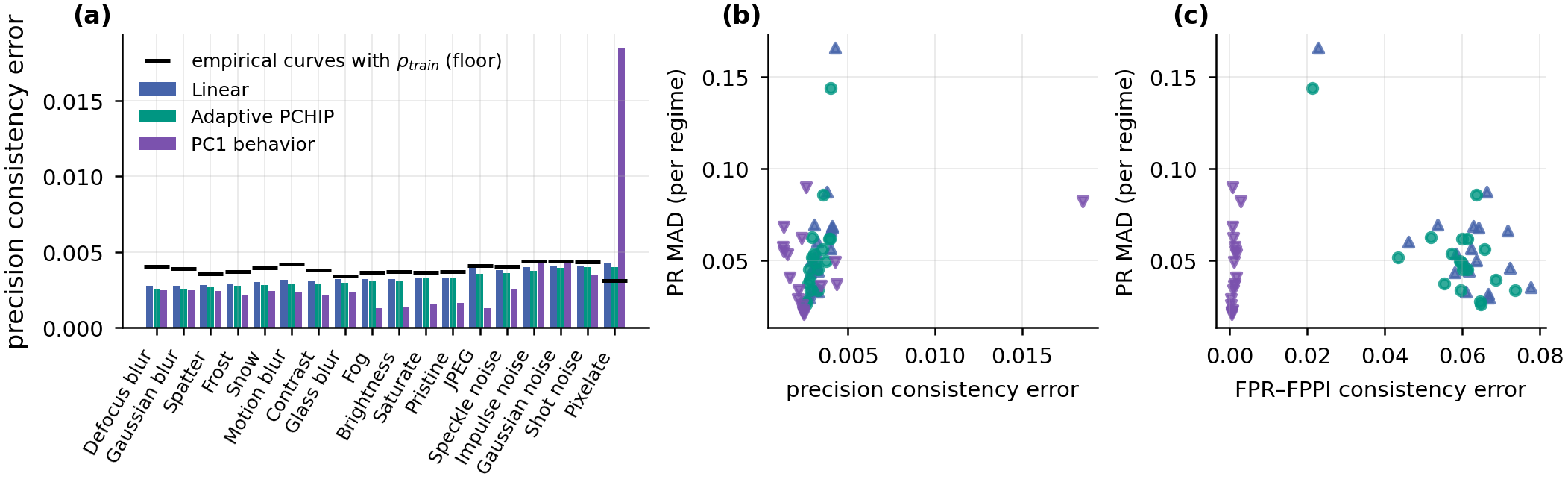}
\caption{Internal consistency diagnostics for predicted behavior components. Predicted precision is largely compatible with TPR/FPPI, whereas separate spline estimates of FPR and FPPI show residual structural inconsistencies.}
\label{S:fig:extended_consistency}
\end{figure}
These diagnostics motivate future work on jointly constrained behavior parameterizations, but do not establish improved operating-point performance from reconstructing precision or adding a consistency loss.

\section{Detector Generalization: Medium YOLO Models}
\label{S:sec:supp_medium}
\paragraph{Setup and protocol.}
The main paper uses YOLOv10-Nano as its controlled proof of concept. We separately repeated the detector-specific feature extraction, monitor training, and final behavior evaluation with fixed pretrained YOLOv10-m and YOLO11-m detectors (person class, $640\times640$ input). For both detectors, hooks capture the three pre-head feature levels (P3/P4/P5, layers 16/19/22) and concatenate global average and maximum pooled activations: 2,304 dimensions for YOLOv10-m and 2,560 for YOLO11-m. These are corresponding multiscale feature levels, not identical feature representations or detection heads. Each detector has separately fitted training targets, feature normalization, and monitors. The original train/validation/test image split and 18 regimes remain fixed (407,472/45,240/85,800 variants); base-image identities are disjoint across splits. We compare the mean-behavior prior, a train-fitted scalar PC1 monitor, and the original 16-uniform-knot PCHIP monitor, with three seeds and 60 epochs for each learned monitor. The selected checkpoint minimizes validation loss. No test-derived PCA, calibration, or monitor selection is used. The detector-level confidence floor is $10^{-3}$ and stored predictions are bounded at 300 detections per image.

\paragraph{Functional behavior reconstruction.}
\Cref{S:tab:medium_curves} reports macro ROC, FROC, and PR curve MAD on all 18 test regimes. Relative to the mean-behavior prior, PCHIP reduces the ROC/FROC/PR errors by approximately 76/81/60\% on YOLOv10-m and 83/85/69\% on YOLO11-m. A train-fitted PC1 monitor also improves substantially over the mean-behavior prior. PCHIP better reconstructs ROC and FROC curves, whereas PC1 has lower geometric PR-curve error on both detectors, concentrated especially in Pixelate. These errors evaluate inference of \emph{condition-level} detector behavior from a single unlabeled image, not intrinsic ground-truth curves for individual images. Cross-seed variability is small; image-level paired uncertainty excludes zero for the improvements over the mean-behavior prior. The PC1 monitor's train-only PCA explains 79.0\% and 75.3\% of its respective behavior representation variance, and cannot be equated with the differently weighted 96.6\% main-paper visualization.

\begin{table}[ht]
\centering
\caption{Medium-detector curve prediction (mean over 18 test regimes). Learned models: mean $\pm$ SD over three seeds. Red. (R/F/P): percentage reduction in ROC/FROC/PR MAD relative to the detector-specific mean-behavior prior. Lower MAD is better.}
\label{S:tab:medium_curves}
\small
\begin{tabular}{llrrrr}
\toprule
Detector & Monitor & ROC MAD $\downarrow$ & FROC MAD $\downarrow$ & PR MAD $\downarrow$ & Red. (R/F/P) \\
\midrule
YOLOv10-m & Mean behavior & 0.0854 & 0.1078 & 0.0966 & -- \\
 & PC1 & 0.0392 $\pm$ 0.0002 & 0.0267 $\pm$ 0.0001 & 0.0297 $\pm$ 0.0000 & 54/75/69 \\
 & PCHIP & 0.0201 $\pm$ 0.0005 & 0.0201 $\pm$ 0.0009 & 0.0390 $\pm$ 0.0009 & 76/81/60 \\
\midrule
YOLO11-m & Mean behavior & 0.1085 & 0.1280 & 0.1163 & -- \\
 & PC1 & 0.0355 $\pm$ 0.0006 & 0.0244 $\pm$ 0.0002 & 0.0291 $\pm$ 0.0001 & 67/81/75 \\
 & PCHIP & 0.0189 $\pm$ 0.0027 & 0.0191 $\pm$ 0.0006 & 0.0362 $\pm$ 0.0010 & 83/85/69 \\
\bottomrule
\end{tabular}

\end{table}

\paragraph{True threshold execution.}
\Cref{S:tab:medium_execution} gives detector performance when each monitor-selected confidence threshold is applied to the \emph{stored detections for that image}. The macro $F_1$ improvement of PCHIP over the mean-behavior prior is $+0.0083$ (95\% paired bootstrap CI $[0.0075,0.0091]$) on YOLOv10-m and $+0.0112$ $([0.0102,0.0122])$ on YOLO11-m; gains over a fixed pristine threshold are $+0.0130$ and $+0.0234$. The incremental gain over PC1 is only $+0.0010$ and $+0.0004$ macro $F_1$, respectively (the latter CI spans zero). On YOLO11-m, however, PCHIP has lower \emph{micro} $F_1$ than PC1 ($0.7477$ versus $0.7598$); the false-positive-heavy Pixelate regime explains the discrepancy. Using each image's threshold instead of the average threshold of its condition changes macro $F_1$ by only about $-0.0001$ for either Medium detector. Gains therefore arise predominantly from inferring a condition-level operating state using single-image evidence, not from within-condition image-specific threshold variation.

\begin{table}[ht]
\centering
\caption{True execution on the full 85,800-variant test set of each Medium detector. $P$ and $R$ denote pooled micro precision and recall; macro $F_1$ averages regimes. All policies use the stored-detection threshold floor $\tau_{\rm exec}=\max(\tau_{\rm pred},0.001)$. Learned models are means over three seeds. $^{\ddagger}$Regime oracle uses test labels and is non-deployable. YOLO11-m PCHIP values are reported although a pre-specified validity check failed (see text).}
\label{S:tab:medium_execution}
\small
\begin{tabular}{llrrrrr}
\toprule
Detector & Policy & Macro $F_1$ & Micro $F_1$ & P & R & FPPI \\
\midrule
YOLOv10-m & Fixed pristine & 0.7478 & 0.7583 & 0.916 & 0.647 & 0.137 \\
 & Mean behavior & 0.7524 & 0.7627 & 0.881 & 0.673 & 0.209 \\
 & PC1 & 0.7597 & 0.7639 & 0.869 & 0.681 & 0.236 \\
 & PCHIP & 0.7608 & 0.7622 & 0.859 & 0.685 & 0.258 \\
 & Regime oracle$^{\ddagger}$ & 0.7615 & 0.7637 & 0.866 & 0.683 & 0.242 \\
\midrule
YOLO11-m & Fixed pristine & 0.7297 & 0.7453 & 0.925 & 0.624 & 0.115 \\
 & Mean behavior & 0.7419 & 0.7570 & 0.869 & 0.670 & 0.231 \\
 & PC1 & 0.7527 & 0.7598 & 0.864 & 0.678 & 0.245 \\
 & PCHIP & 0.7531 & 0.7477 & 0.823 & 0.685 & 0.337 \\
 & Regime oracle$^{\ddagger}$ & 0.7556 & 0.7580 & 0.853 & 0.682 & 0.270 \\
\bottomrule
\end{tabular}

\end{table}

\paragraph{FPPI-constrained policies.}
The same predicted threshold functions support validation-calibrated recall maximization under an FPPI budget $B$. \Cref{S:tab:medium_fppi} contrasts PC1 and PCHIP under the original validation-only calibration rule. PCHIP obtains greater recall across the shown budgets, for example an absolute gain of 0.024/0.021 at $B=0.1$ on YOLOv10-m/YOLO11-m, respectively. This does \emph{not} imply guaranteed constraint satisfaction: at $B=0.2$, PCHIP exceeds the nominal budget in 4.3 and 5.7 of 18 test conditions on average (across seeds), versus 1.0 and 0.0 for PC1; test-regime violations, including the 10\% relative-tolerance diagnostic, are shown explicitly. Uncalibrated policies frequently violate the budget, and were not used to establish safe operation.

\begin{table}[ht]
\centering
\caption{Validation-calibrated FPPI policies. Recall is macro over 18 regimes; violations are the mean number of test regimes exceeding nominal budget $B$ (and a 10\% larger budget) across three seeds. Higher recall does not establish a hard FPPI guarantee.}
\label{S:tab:medium_fppi}
\small
\begin{tabular}{lllrrr}
\toprule
Detector & Budget $B$ & Policy & Recall $\uparrow$ & Violations & Violations (+10\%) \\
\midrule
YOLOv10-m & 0.05 & PC1 & 0.529 & 0.0 & 0.0 \\
 &  & PCHIP & 0.558 & 0.0 & 0.0 \\
 & 0.1 & PC1 & 0.603 & 0.0 & 0.0 \\
 &  & PCHIP & 0.627 & 1.0 & 0.0 \\
 & 0.2 & PC1 & 0.657 & 1.0 & 0.0 \\
 &  & PCHIP & 0.677 & 4.3 & 0.3 \\
 & 0.5 & PC1 & 0.713 & 2.0 & 0.0 \\
 &  & PCHIP & 0.726 & 1.0 & 0.0 \\
\midrule
YOLO11-m & 0.05 & PC1 & 0.518 & 0.0 & 0.0 \\
 &  & PCHIP & 0.538 & 0.0 & 0.0 \\
 & 0.1 & PC1 & 0.585 & 0.0 & 0.0 \\
 &  & PCHIP & 0.607 & 0.0 & 0.0 \\
 & 0.2 & PC1 & 0.652 & 0.0 & 0.0 \\
 &  & PCHIP & 0.670 & 5.7 & 1.3 \\
 & 0.5 & PC1 & 0.707 & 1.0 & 0.0 \\
 &  & PCHIP & 0.723 & 2.7 & 0.0 \\
\bottomrule
\end{tabular}

\end{table}

\paragraph{Observed floor-related failure and protocol transparency.}
An independent recount reconstructed all saved TP/FP/FN counts exactly from the stored detections and confirmed that every result uses $\tau_{\rm exec}=\max(\tau_{\rm pred},0.001)$. Thus clipping is an appropriate, uniform implementation of the \emph{stored-detector pipeline}; it does not recover hypothetical discarded detections below that floor. Threshold selection first maximizes predicted $F_1$ over the full $[0,1]$ grid and then clips for execution. This rule does not itself cause the low-threshold error: on available YOLO11-m Pixelate examples a constrained argmax $\tau\geq0.001$ still chooses $0.001$.

For YOLO11-m Pixelate (severities 3--5), PCHIP chooses $\tau_{\rm pred}=0$ for about 99.3--99.6\% of variants in each seed (4,915--4,931 of 4,950). This \emph{failed} the pre-specified score-floor validity check (at most 0.1\% of variants globally and 1\% per condition). A protocol revision, fixed before any further inspection of test performance, allowed reporting the full-population operating results with clipping and this explicit failure disclosure; \emph{the check remains failed, and the YOLO11-m PCHIP operating results are reported under the revised protocol}. In this regime, executed PCHIP reaches $F_1\simeq0.285$ but incurs about 8,954 false positives (FPPI $\simeq1.81$); the regime oracle reaches $F_1\simeq0.316$. In train and validation the optimal confidence threshold is near $0.0044$/$0.0049$, whereas the original PCHIP representation's first interpolation interval spans $[0,0.067]$. This is consistent with insufficient low-threshold resolution, but not a causal knot-density ablation. A separate, \emph{post-hoc} paired sensitivity comparison dropping the entire Pixelate regime from both detectors (80,850 variants each) removes YOLO11-m PCHIP's micro-$F_1$ deficit; it is not a replacement for the full-population result. YOLOv10-m has only two clipped test variants, and its original base-image-exclusion sensitivity changes evaluated metrics by at most $10^{-4}$.

\paragraph{Detection-cap limitation and scope.}
The acquisition limit of 300 stored detections per image was reached on three YOLOv10-m and 2,180 YOLO11-m images, primarily in noise regimes. At affected very low thresholds, empirical false-positive counts can be censored; precision and $F_1$ should not be interpreted as lower bounds for a hypothetical uncapped pipeline. The YOLO11-m Pixelate failure is not caused by that cap (no affected Pixelate test or validation image reached it). The empirical transfer shown here covers two further fixed, pretrained, COCO-derived person detectors, not arbitrary architecture/shift generalization. The richer PCHIP representation provides better recall/false-positive behavior reconstruction but does not guarantee a better operating point; for YOLO11-m, nonuniform or additional knots near the confidence floor remain a \emph{prospective}, not evaluated, extension. For YOLOv10-Nano, representation floors of alternative knot layouts are reported in \Cref{S:sec:followup}.

\section{Follow-Up Studies: Functional versus Discrete Realizations}
\label{S:sec:followup}
The main benchmark pools severities 3--5 into 18 regimes, so every image inherits one of 18 discrete targets. The follow-up studies below test what this benchmark can and cannot show about functional inference; they answer RQ4 of the main paper, whose Table~3 summarizes \Cref{S:tab:fu_regimes,S:tab:fu_interp}. All use the frozen YOLOv10-Nano detector, the stored features and detections, and the original train/validation/test split. They were run after the main study and reuse its test pool.

\paragraph{Studies.} \emph{Pre-specified} means that hypotheses, endpoints, and decision rules were fixed in a time-stamped protocol before any training or test evaluation of that study.
\begin{itemize}
\item \textbf{Severity interpolation, lookup controls, score-only monitor:} pre-specified, with two protocol revisions recorded before the affected runs; exploratory in motivation and evaluated on the main test pool.
\item \textbf{Diagnostics} (representation floors, confusion analysis, mixture weights, knot layouts): not pre-specified; the knot layout is chosen from training targets only, by a rule fixed before the computation.
\item \textbf{Regime-label-free decoders} (PCHIP-32, template decoder): pre-specified; exploratory on the main test pool.
\item \textbf{Confirmation test:} pre-specified test of the one positive interpolation finding of the regime-label-free decoders. Severity 2 had not been evaluated before; the test base images are the same.
\end{itemize}

\paragraph{Common protocol.} Learned arms follow the main protocol (two-layer readout on the same pooled features, AdamW, 60 epochs, checkpoint with minimum validation loss, three seeds). Execution applies $\tau_{\rm exec}=\max(\tau_{\rm pred},10^{-3})$; $\tau_{\rm pred}=0$ never occurred in any arm, condition, or seed. \emph{Mean MAD} is the average of ROC, FROC, and PR MAD. Confidence intervals are paired percentile bootstraps over base images (2,000 resamples, seed 12345); learned arms are seed-averaged within each replicate. $\Delta$ denotes the first-named minus the second-named arm. Our evaluation code for these studies reproduces the main execution table bitwise and the main MAD values within $2.5\times10^{-6}$. \emph{Caveat:} the intervals cover image sampling only. The seed SD of the lookups (up to 0.0019 mean MAD) is comparable to several of the reported contrasts.

\paragraph{Arms.}
\begin{itemize}
\item \textbf{Hard lookup$^\dagger$:} a classifier with the same readout capacity, trained with cross-entropy over the training conditions, returns the stored training-pool curve and the validation-optimal threshold of the predicted condition.
\item \textbf{Soft lookup$^\dagger$:} mixes the stored TPR, FPR, and FPPI curves with the class probabilities and recomputes precision as $\mathrm{TPR}\cdot\rho/(\mathrm{TPR}\cdot\rho+\mathrm{FPPI})$, with $\rho=2.189$ annotated persons per image; the threshold maximizes $F_1$ on the mixture.
\item \textbf{Score-only:} the PCHIP-16 decoder on 41 fixed statistics of the person-detection scores (histogram, counts above fixed score levels, top-10 scores, log detection count).
\item \textbf{PCHIP-32:} the main decoder with 32 uniform knots, selected from the knot-layout diagnostic (\Cref{S:tab:fu_knots}).
\item \textbf{Template decoder:} a softmax ($T=1$, no entropy term) over the stored training-pool curves, trained with the curve loss only; no regime label is a training target.
\end{itemize}
$^\dagger$Regime-supervised controls: they use regime labels as a training target and are therefore not DOBI realizations.

\subsection{Pooled 18-Regime Benchmark}
\begin{table}[H]
\centering
\caption{18 test regimes: curve MAD and true-execution $F_1$. Learned arms: seed mean ($\pm$ SD for mean MAD). $^\ddagger$Non-deployable; knows the true regime.}
\label{S:tab:fu_regimes}
\small
\begin{tabular}{lcccccc}
\toprule
Arm & Mean MAD & ROC & FROC & PR & Macro $F_1$ & Micro $F_1$ \\
\midrule
Mean behavior & 0.1214 & 0.1077 & 0.1287 & 0.1278 & 0.6536 & 0.6617 \\
Score-only PCHIP-16 & 0.1087$\pm$0.0005 & 0.0933 & 0.1127 & 0.1201 & 0.6574 & 0.6643 \\
PCHIP-16 (main paper) & 0.0437$\pm$0.0006 & 0.0433 & 0.0342 & 0.0536 & 0.6604 & 0.6622 \\
PCHIP-32 & 0.0362$\pm$0.0006 & 0.0358 & 0.0336 & 0.0392 & 0.6596 & 0.6615 \\
Template decoder & 0.0291$\pm$0.0001 & 0.0253 & 0.0313 & 0.0307 & 0.6605 & 0.6620 \\
Hard lookup$^\dagger$ & 0.0245$\pm$0.0006 & 0.0196 & 0.0280 & 0.0259 & 0.6596 & 0.6605 \\
Soft lookup$^\dagger$ & 0.0256$\pm$0.0008 & 0.0207 & 0.0289 & 0.0271 & 0.6608 & 0.6625 \\
Regime-ID$^\ddagger$ & -- & -- & -- & -- & 0.6604 & 0.6605 \\
\bottomrule
\end{tabular}

\end{table}
Both lookups fit the curves better than the main PCHIP-16 monitor in every regime, although the 18-way classifier is correct for only $75.9\pm0.6\%$ of test images. Decision quality is equivalent within 0.002 macro $F_1$ for all deployable arms and Regime-ID. In curve MAD, PCHIP-16 is worse than the mean-behavior prior for 4 of 18 regimes (contrast, motion blur, snow, speckle noise), the lookups for 2 (snow, speckle noise). The template decoder beats the soft lookup in 6 of 18 regimes (Gaussian, impulse, glass blur, motion blur, pixelate, spatter); its average deficit is concentrated in JPEG compression (0.0644 versus 0.0185). Score-only reaches 0.1087, only modestly better than the mean-behavior prior. Its micro $F_1$ is nevertheless higher than PCHIP-16's (0.6643 versus 0.6622) because it selects more conservative thresholds. The 300-detection cap (12.1\% of test variants) does not explain its weakness: its MAD is 0.1109 at the cap and 0.1094 below it. Only this fixed 41-dimensional summary was tested, not every possible use of the detector output.

\paragraph{Representation versus inference.} \Cref{S:tab:fu_decomp} compares each arm with the best fit its representation allows. On the 18 regimes, the PCHIP-16 floor (0.0190) is about as large as PCHIP-16's whole gap to the hard lookup (0.0192); about 58\% of the gap is representation excess over the stored-curve floor (0.0080), mainly in ROC and PR and worst for pixelate and the noise types. At unseen severity 4, inference error dominates: even the best stored training curve would reach about 0.020, whereas all deployable arms remain at 0.078--0.099. The floors are fitted with the training loss rather than MAD.
\begin{table}[H]
\centering
\caption{Representation floors: best achievable MAD of each representation against the test targets, versus the trained result. \emph{Lookup floor}: stored training curve of the correct condition.}
\label{S:tab:fu_decomp}
\small
\begin{tabular}{llccccc}
\toprule
Setup & Curve & PCHIP-16 floor & PCHIP-16 & Lookup floor & Hard lookup & Soft lookup \\
\midrule
18 regimes & ROC & 0.0286 & 0.0433 & 0.0055 & 0.0196 & 0.0207 \\
 & FROC & 0.0027 & 0.0342 & 0.0103 & 0.0280 & 0.0289 \\
 & PR & 0.0256 & 0.0536 & 0.0081 & 0.0259 & 0.0271 \\
 & Mean & 0.0190 & 0.0437 & 0.0080 & 0.0245 & 0.0256 \\
\midrule
sev.\,4 & ROC & 0.0317 & 0.0721 & 0.0066 & 0.0855 & 0.0689 \\
 & FROC & 0.0030 & 0.0836 & 0.0099 & 0.1039 & 0.0837 \\
 & PR & 0.0296 & 0.0992 & 0.0087 & 0.1070 & 0.0861 \\
 & Mean & 0.0215 & 0.0850 & 0.0084 & 0.0988 & 0.0795 \\
\bottomrule
\end{tabular}

\end{table}

\paragraph{Confusions.} Misclassified images receive a lookup curve with MAD 0.077, against 0.182 for a uniformly drawn wrong regime (0.42$\times$). The rank relation between confusion rate and standardized behavior distance is weak (Spearman $-0.113$ [$-0.152$, $-0.101$]); 27\% of regime pairs are never confused, and among the others $\rho=-0.37$ (descriptive). The most frequent confusion of a regime is its nearest regime in behavior space in only 3 of 18 cases. Feature similarity and behavioral similarity are therefore related but not identical.

\paragraph{Knot layouts.} \Cref{S:tab:fu_knots} lists representation floors on the 18 \emph{training} regimes. Additional uniform knots reduce the floor in every regime, most for pixelate (0.059 to 0.022) and the noise types. Redistributing 16 knots only moves the error between curves: all-score quantiles (identical to the main-paper ablation) improve ROC but degrade PR, and log-spaced knots give the best ROC and FROC floors but a poor PR floor.
\begin{table}[H]
\centering
\caption{Representation floors of knot layouts on the 18 training regimes (training targets only). The selection rule (smallest mean floor; ties within 0.001 resolved toward fewer, then uniform knots) was fixed before computation.}
\label{S:tab:fu_knots}
\small
\begin{tabular}{lccccc}
\toprule
Representation & Knots & ROC & FROC & PR & Mean \\
\midrule
uniform (main paper) & 16 & 0.0280 & 0.0023 & 0.0251 & 0.0185 \\
uniform & 24 & 0.0167 & 0.0024 & 0.0138 & 0.0110 \\
uniform (selected) & 32 & 0.0122 & 0.0017 & 0.0089 & 0.0076 \\
all-score quantiles & 16 & 0.0099 & 0.0359 & 0.0732 & 0.0397 \\
log-spaced & 16 & 0.0050 & 0.0010 & 0.0589 & 0.0216 \\
\bottomrule
\end{tabular}

\end{table}

\subsection{Unseen Intermediate Severities}
\begin{table}[H]
\centering
\caption{Interpolation to unseen severities. Left: trained on pristine and severities $\{3,5\}$, tested on severity 4 (17 conditions). Right: pre-specified confirmation test, trained on pristine and $\{1,3,5\}$, tested on $\{2,4\}$ (34 conditions); the template decoder was not part of it. $^\ddagger$Non-deployable: mean of the true training curves of severities 3 and 5 of the correct type. $^\ast$Upper reference trained with severity 4 seen.}
\label{S:tab:fu_interp}
\resizebox{\textwidth}{!}{%
\begin{tabular}{lcccccc}
\toprule
 & \multicolumn{2}{c}{Train $\{3,5\}$, test 4} & \multicolumn{4}{c}{Train $\{1,3,5\}$, test $\{2,4\}$ (pre-specified)} \\
\cmidrule(lr){2-3}\cmidrule(lr){4-7}
Arm & Mean MAD & Macro $F_1$ & Mean MAD & sev.\,2 & sev.\,4 & Macro $F_1$ \\
\midrule
Mean behavior & 0.1275 & 0.6346 & 0.1096 & 0.0907 & 0.1284 & 0.7124 \\
PCHIP-16 & 0.0850$\pm$0.0004 & 0.6416 & 0.0597$\pm$0.0004 & 0.0345 & 0.0850 & 0.7193 \\
PCHIP-32 & 0.0783$\pm$0.0005 & 0.6429 & 0.0557$\pm$0.0002 & 0.0320 & 0.0794 & 0.7197 \\
Template decoder & 0.0846$\pm$0.0006 & 0.6417 & -- & -- & -- & -- \\
Hard lookup$^\dagger$ & 0.0988$\pm$0.0015 & 0.6410 & 0.0705$\pm$0.0013 & 0.0411 & 0.0998 & 0.7179 \\
Soft lookup$^\dagger$ & 0.0795$\pm$0.0012 & 0.6427 & 0.0549$\pm$0.0019 & 0.0330 & 0.0767 & 0.7198 \\
\midrule
Neighbor oracle$^\ddagger$ & 0.0312 & 0.6439 & -- & -- & -- & -- \\
PCHIP-16, sev.\,4 seen$^\ast$ & 0.0534$\pm$0.0007 & 0.6453 & -- & -- & -- & -- \\
\bottomrule
\end{tabular}
}
\end{table}
In the first setting, PCHIP-32 was the only arm better than the soft lookup ($\Delta$ $-0.0012$ [$-0.0016$, $-0.0008$]). Because this finding was exploratory, it was then tested with a pre-specified hypothesis (PCHIP-32 reconstructs unseen intermediate severities better than the soft lookup) and decision rule (pooled $\Delta$ mean MAD with CI entirely below zero). The hypothesis was \textbf{refuted}: $\Delta=+0.0008$ [$+0.0005$, $+0.0011$] (\Cref{S:tab:fu_contrasts}). The direction differs by severity: PCHIP-32 is better at severity 2 (in 11 of 17 types) but worse at severity 4 (better in only 3 of 17 types), where it loses mainly on smooth photometric types (brightness, fog, saturate, frost). Adding severity 1 to training improved the soft lookup at severity 4 (0.0767 versus 0.0795) but not PCHIP-32 (0.0794 versus 0.0783). Decisions are practically identical ($\Delta$ macro $F_1$ $-0.0002$ [$-0.0007$, $+0.0003$]). On these unseen conditions, each learned arm is worse than the mean-behavior prior for 9--12 of 34 conditions (depending on the arm), most strongly for contrast at severity 4 (prior 0.026 versus 0.129 for PCHIP-32).

Without a classification objective, the template decoder does not collapse (mean normalized weight entropy 0.12 on the 18 regimes and 0.20 in the severity-4 setting), but it places only about 43\% of its mass on the same degradation type at severity 4, against 73\% for the soft lookup: it selects templates by curve similarity rather than degradation identity, which does not help interpolation here.

\begin{table}[H]
\centering
\caption{Paired contrasts ($\Delta$ = first minus second arm; negative $\Delta$ MAD and positive $\Delta F_1$ favor the first arm).}
\label{S:tab:fu_contrasts}
\resizebox{\textwidth}{!}{%
\begin{tabular}{llll}
\toprule
Contrast & Conditions & $\Delta$ mean MAD [95\% CI] & $\Delta$ macro $F_1$ [95\% CI] \\
\midrule
PCHIP-32 $-$ soft lookup & sev.\,2+4 (primary) & +0.0008 [+0.0005, +0.0011] & -0.0002 [-0.0007, +0.0003] \\
 & sev.\,2 & -0.0010 [-0.0014, -0.0006] & -0.0002 [-0.0007, +0.0003] \\
 & sev.\,4 & +0.0026 [+0.0022, +0.0030] & -0.0001 [-0.0009, +0.0007] \\
PCHIP-32 $-$ hard lookup & sev.\,2+4 & -0.0148 [-0.0151, -0.0144] & +0.0017 [+0.0011, +0.0024] \\
 & sev.\,2 & -0.0091 [-0.0095, -0.0086] & +0.0011 [+0.0005, +0.0018] \\
 & sev.\,4 & -0.0205 [-0.0210, -0.0200] & +0.0023 [+0.0013, +0.0034] \\
PCHIP-32 $-$ PCHIP-16 & sev.\,2+4 & -0.0041 [-0.0042, -0.0039] & +0.0003 [-0.0001, +0.0007] \\
 & sev.\,2 & -0.0025 [-0.0027, -0.0023] & -0.0002 [-0.0006, +0.0002] \\
 & sev.\,4 & -0.0057 [-0.0059, -0.0054] & +0.0009 [+0.0001, +0.0016] \\
Template $-$ soft lookup & 18 regimes & +0.0035 [+0.0032, +0.0039] & -0.0004 [-0.0006, -0.0001] \\
PCHIP-32 $-$ soft lookup & 18 regimes & +0.0107 [+0.0103, +0.0111] & -0.0012 [-0.0016, -0.0008] \\
Template $-$ soft lookup & sev.\,4, train $\{3,5\}$ & +0.0052 [+0.0048, +0.0055] & -0.0010 [-0.0017, -0.0003] \\
PCHIP-32 $-$ soft lookup & sev.\,4, train $\{3,5\}$ & -0.0012 [-0.0016, -0.0008] & +0.0003 [-0.0006, +0.0011] \\
\bottomrule
\end{tabular}
}
\end{table}

\paragraph{Protocol deviation in the confirmation test.} A pre-specified feature-equivalence check failed: 1,920 of 2,000 re-extracted severity-3 feature vectors matched the stored features bitwise, but the 80 vectors of the final partial batch deviated (relative L2 at most $1.04\times10^{-3}$), consistent with batch-shape-dependent cuDNN numerics. The check is reported as failed. A protocol revision before any training padded all batches to full size; the repeated check on a new sample matched 2,000 of 2,000 vectors bitwise. Hypothesis, endpoint, and training set were unchanged. The partial-batch rows of the original stored features (48/56/40 rows of train/validation/test) differ by at most about 0.1\% relative L2 from full-batch features and are used unchanged, as in all earlier experiments.

\paragraph{Summary.} Regime-supervised lookups fit the pooled regimes best. Regime-label-free decoders approach them, and part of the original gap is a representation limit of 16 uniform knots. Graded inference beats the hard lookup throughout. A pre-specified test found no advantage of the continuous decoder over the soft lookup on unseen intermediate severities, and decisions are insensitive to all of these differences. The benchmark thus shows that regime-level behavior is recoverable without regime labels, but not that continuous representations outperform discrete ones. \Cref{S:sec:continuous_supp} tests where they do, under continuous and compositional probing.

\clearpage
\section{Continuous and Compositional Probing (RQ5)}
\label{S:sec:continuous_supp}
The discrete benchmark of the main study assigns every image one of 18 regime targets; there, a soft lookup over stored curves is the natural estimator of the posterior mixture (main paper, Section~3.3). This section documents two experiments in which conditions are drawn from a continuous space, so that test conditions can lie between stored conditions or combine degradations that never co-occur in training.

\paragraph{Experiments.} \emph{Initial experiment:} monitors are trained and evaluated on 142 test conditions that do not occur in training or validation. \emph{Confirmation experiment:} the trained monitors are frozen and evaluated on 710 freshly drawn test conditions. Both were pre-specified: hypotheses, endpoints, and decision rules were fixed in time-stamped protocols before any image of the respective experiment was generated. A third, descriptive analysis repeats the detector inference of the initial evaluation with FP32 matrix multiplication.

\subsection{Design}
\paragraph{Detector, data, and targets.} The frozen YOLOv10-Nano of the main study (input 640, letterbox, fp32, fixed batches of 128 with padding, confidence floor 0.001, NMS IoU 0.7, at most 300 detections), class person, pooled P3--P5 neck features (896-dimensional) as monitor input. Base images: the person subset of COCO with the fixed splits of the main study (7{,}836 training, 870 validation, 1{,}650 test). For a group of images, all person detections are pooled into one operating sweep (greedy matching at IoU $\geq$ 0.5); TPR, the candidate-normalized FPR, FPPI, and precision are computed on 201 thresholds and converted to ROC, FROC, and PR curves. Each condition is evaluated on its own group of base images, whose pooled curve is the condition's target curve.

\paragraph{Condition generator.} Strength $s$ is real-valued on the 1--5 scale of the common-corruptions parameter tables; parameters are interpolated component-wise between neighboring integer levels and coincide with the tables at integer $s$. A composition applies one degradation, quantizes to 8 bits, and applies the second, in a fixed order derived from image formation (weather/medium, exposure/color, optics, sensor noise, digital). All random draws are seeded from a hash of condition, base image, and processing step, so every image is reproducible bit for bit.

\paragraph{Degradation types.} Of 17 candidates, 15 were admitted. Glass blur was excluded in advance (integer, non-monotone parameters). Pixelate and snow failed a pre-specified monotonicity check (dissimilarity to the pristine image non-decreasing in $s$ within 2\% of its range, 50 images and common random numbers per type). Saturate is restricted to $s\in[3,5]$ (over-saturation only), and spatter is split into water ($s\in[1.5,3]$) and mud ($s\in[4,5]$), which use different mechanisms. This yields 104 unordered type pairs.

\paragraph{Condition sets.} In the initial experiment, each type's range is divided into 14 cells; four non-adjacent interior cells are test cells, and one training strength is drawn in the middle half of each other cell (10 per type), so that interpolation test strengths are at least $0.75$ cell widths from all training strengths. Of the 104 pairs, 26 are held out as \emph{unseen pairs}. The confirmation experiment draws fresh conditions with a new seed and distance rules to all earlier conditions.

\begin{table}[H]
\centering
\caption{Condition sets. Training conditions also form the soft lookup's curve bank.}
\label{S:tab:cont_sets}
\small
\begin{tabular}{lll}
\toprule
Experiment & Set & Conditions \\
\midrule
Both & Training: 15 types $\times$ 10 strengths, 78 seen pairs $\times$ 2 combinations, pristine & 307 \\
 & Validation: 15 types $\times$ 2 strengths, 40 combinations of training pairs & 70 \\
\midrule
Initial & Test: interpolation (cell midpoints) & 60 \\
 & Test: compositions of unseen pairs (26 pairs $\times$ 2) & 52 \\
 & Test: compositions of seen pairs (new strength combinations) & 30 \\
 & \textbf{Endpoint set: all unseen test conditions} & \textbf{142} \\
\midrule
Confirmation & Interpolation (one new strength per test cell) & 60 \\
 & Compositions of unseen pairs (26 pairs $\times$ 16) & 416 \\
 & Compositions of seen pairs (78 pairs $\times$ 3) & 234 \\
 & \textbf{All compositions (hypotheses 1 and 3)} & \textbf{650} \\
\bottomrule
\end{tabular}
\end{table}

\paragraph{Group size.} Every training and test condition uses $n=1{,}200$ base images (validation 600). $n$ was chosen on training data only as the smallest candidate whose median target-curve noise (mean curve distance between a random subgroup and the full curve) is at most 0.010, about a quarter of the main study's typical curve error. The initial candidate set up to 800 did not reach the threshold and was extended before any test data existed.

\begin{table}[H]
\centering
\caption{Median target-curve noise by group size (training conditions only).}
\label{S:tab:cont_n}
\small
\begin{tabular}{lcccccc}
\toprule
$n$ & 200 & 300 & 500 & 800 & 1{,}000 & 1{,}200 \\
Median noise & 0.0240 & 0.0187 & 0.0147 & 0.0115 & 0.0100 & \textbf{0.0092} \\
\bottomrule
\end{tabular}
\end{table}

\paragraph{Monitors.} All learned monitors use an MLP ($896\to1024\to1024$), AdamW, 60 epochs, and deterministic training; curve monitors are trained with Smooth-$L_1$ on TPR, FPR, $\log(1+\mathrm{FPPI})$, and precision, and the checkpoint with the lowest validation loss is selected. The \emph{functional decoder} (primary, 5 seeds) predicts monotone control points with PCHIP on 32 uniform knots and uses no condition labels. The \emph{soft lookup} (5 seeds) is a classifier over the 307 training conditions whose output mixes the stored training curves with the class probabilities (precision recomputed from the mixed TPR and FPPI); its checkpoint is selected by the curve loss of the mixture on the validation conditions, the most favorable fair selection for the lookup. Further monitors: hard lookup (arg-max class), a 16-knot decoder with the knot placement of the main study, the mean-behavior prior, brightness and score-only baselines, and, in the initial experiment, a non-deployable oracle replicate (the same condition on different base images). The confirmation experiment reuses all 19 trained models unchanged.

\paragraph{Metric and statistics.} Per image, the mean absolute distance between predicted and target curve is averaged over ROC, FROC, and PR ($M$), averaged over a condition's images, and macro-averaged over conditions. $\Delta=M(\text{decoder})-M(\text{soft lookup})$; $\Delta_c$ is the same difference on one condition. Intervals are paired bootstraps over the test base images (2{,}000 resamples) in which target curves are recomputed in every resample; seed means are formed inside each resample. The seed SD is the spread of the per-seed estimates. Distance $d(c)$ is the Euclidean distance, in normalized severity coordinates ($s/5$ per type, 0 if absent), to the nearest of the 307 training conditions; it depends only on the design and was frozen before any image existed.

\paragraph{Decision rules.} \emph{Initial experiment:} single endpoint $\Delta$ over all 142 test conditions; met if the upper percentile limit is below 0 and $|\hat\Delta|$ exceeds the lookup's seed SD. \emph{Confirmation experiment:} three hypotheses in fixed sequence (a hypothesis counts as confirmatory only if all preceding ones are supported). Hypotheses 1 and 2 are supported if both the percentile and the basic interval exclude 0 on the predicted side and $|\hat\Delta|$ exceeds the seed SD of both monitors; hypothesis 3 if both intervals of Spearman's $\rho$ lie below 0. The number of conditions was chosen for a minimum detectable effect of about 0.003 (80\% power).

\subsection{Results}
\paragraph{Initial experiment.} The single endpoint was met but small ($\Delta=-0.00065$, $|\hat\Delta|$ 1.45 times the lookup's seed SD), and it depends on the interval method (\Cref{S:tab:cont_initial}): recomputing targets in each resample shifts the replicate distribution, and the basic interval contains 0. The subsets show opposing effects: the lookup is better for interpolation, the decoder for compositions. The overall value is their weighted difference (42\% interpolation conditions). This motivated the confirmation experiment.

\begin{table}[H]
\centering
\caption{Initial experiment: functional decoder (32 knots) minus soft lookup, all interval variants. Fixed targets: bootstrap without recomputing target curves (descriptive).}
\label{S:tab:cont_initial}
\small
\setlength{\tabcolsep}{3pt}
\begin{tabular}{lccccc}
\toprule
Set & $\Delta$ & Percentile & Basic & BCa & Fixed targets (pct.) \\
\midrule
All unseen (endpoint) & $-0.0007$ & [$-0.0015$, $-0.0004$] & [$-0.0009$, $+0.0002$] & [$-0.0009$, $-0.0000$] & [$-0.0010$, $-0.0004$] \\
Interpolation & $+0.0067$ & [$+0.0058$, $+0.0069$] & [$+0.0066$, $+0.0077$] & [$+0.0065$, $+0.0073$] & [$+0.0064$, $+0.0071$] \\
Unseen pairs & $-0.0067$ & [$-0.0082$, $-0.0059$] & [$-0.0075$, $-0.0052$] & [$-0.0078$, $-0.0056$] & [$-0.0071$, $-0.0063$] \\
Seen pairs & $-0.0050$ & [$-0.0060$, $-0.0043$] & [$-0.0056$, $-0.0039$] & [$-0.0057$, $-0.0040$] & [$-0.0054$, $-0.0045$] \\
\bottomrule
\end{tabular}

\end{table}

\begin{table}[H]
\centering
\caption{Initial experiment: mean curve error $M$ per monitor and set (seed mean $\pm$ seed SD). $^\dagger$Trained with condition labels. $^\ddagger$Not deployable.}
\label{S:tab:cont_arms_initial}
\small
\setlength{\tabcolsep}{3pt}
\begin{tabular}{lcccc}
\toprule
Monitor & All unseen (142) & Interpolation (60) & Unseen pairs (52) & Seen pairs (30) \\
\midrule
Functional decoder, 32 knots & 0.0688$\pm$0.0006 & 0.0458$\pm$0.0014 & 0.0854$\pm$0.0018 & 0.0860$\pm$0.0015 \\
Soft lookup$^\dagger$ & 0.0694$\pm$0.0004 & 0.0390$\pm$0.0011 & 0.0921$\pm$0.0020 & 0.0909$\pm$0.0010 \\
Hard lookup$^\dagger$ & 0.0800$\pm$0.0011 & 0.0505$\pm$0.0021 & 0.1014$\pm$0.0017 & 0.1019$\pm$0.0012 \\
Functional decoder, 16 knots & 0.0748$\pm$0.0012 & 0.0515$\pm$0.0022 & 0.0920$\pm$0.0019 & 0.0917$\pm$0.0020 \\
Mean behavior & 0.1878 & 0.1586 & 0.2092 & 0.2091 \\
Brightness & 0.1687$\pm$0.0021 & 0.1094$\pm$0.0028 & 0.2119$\pm$0.0030 & 0.2126$\pm$0.0007 \\
Score only & 0.1267$\pm$0.0018 & 0.1118$\pm$0.0027 & 0.1342$\pm$0.0014 & 0.1436$\pm$0.0008 \\
Oracle replicate$^\ddagger$ & 0.0152 & 0.0128 & 0.0195 & 0.0127 \\
\bottomrule
\end{tabular}

\end{table}

\paragraph{Confirmation experiment.} All three hypotheses are supported (\Cref{S:tab:cont_hyp}). Every interval variant, with recomputed or fixed targets, excludes 0 on the same side for all compositions, both composition subsets, and interpolation (\Cref{S:tab:cont_bysets}).

\begin{table}[H]
\centering
\caption{Confirmation experiment: pre-specified hypotheses (fixed sequence).}
\label{S:tab:cont_hyp}
\small
\begin{tabular}{llcccc}
\toprule
Order & Hypothesis & Estimate & Percentile 95\% CI & Basic 95\% CI & Result \\
\midrule
1 & Compositions: decoder better ($\Delta<0$) & $-0.0043$ & [$-0.0050$, $-0.0037$] & [$-0.0049$, $-0.0037$] & supported \\
2 & Interpolation: lookup better ($\Delta>0$) & $+0.0067$ & [$+0.0058$, $+0.0070$] & [$+0.0065$, $+0.0076$] & supported \\
3 & Distance: $\rho(d,\Delta_c)<0$ & $-0.180$ & [$-0.258$, $-0.102$] & [$-0.258$, $-0.102$] & supported \\
\bottomrule
\end{tabular}

\end{table}

\begin{table}[H]
\centering
\caption{Confirmation experiment: $\Delta$ by set with all interval variants (top), and the typical case (bottom): median $\Delta_c$, share of conditions where the decoder is better, and seed SDs of decoder and lookup. Basic intervals of shares near 0 can extend beyond $[0,1]$ (interpolation).}
\label{S:tab:cont_bysets}
\resizebox{\textwidth}{!}{%
\setlength{\tabcolsep}{3pt}
\begin{tabular}{lrccccc}
\toprule
Set & $n$ & $\Delta$ & Percentile & Basic & BCa & Fixed targets (pct.) \\
\midrule
All test conditions & 710 & $-0.0034$ & [$-0.0041$, $-0.0029$] & [$-0.0039$, $-0.0028$] & [$-0.0040$, $-0.0029$] & [$-0.0037$, $-0.0031$] \\
All compositions & 650 & $-0.0043$ & [$-0.0050$, $-0.0037$] & [$-0.0049$, $-0.0037$] & [$-0.0050$, $-0.0038$] & [$-0.0046$, $-0.0041$] \\
Unseen pairs & 416 & $-0.0063$ & [$-0.0071$, $-0.0056$] & [$-0.0071$, $-0.0055$] & [$-0.0072$, $-0.0057$] & [$-0.0066$, $-0.0060$] \\
Seen pairs & 234 & $-0.0008$ & [$-0.0014$, $-0.0003$] & [$-0.0012$, $-0.0002$] & [$-0.0013$, $-0.0002$] & [$-0.0011$, $-0.0005$] \\
Interpolation & 60 & $+0.0067$ & [$+0.0058$, $+0.0070$] & [$+0.0065$, $+0.0076$] & [$+0.0065$, $+0.0072$] & [$+0.0063$, $+0.0071$] \\
\midrule
 & & Median $\Delta_c$ & Median, basic CI & Decoder better & Share, basic CI & SD$_\text{seed}$ dec./lookup \\
\midrule
All test conditions & 710 & $+0.0024$ & [$+0.0019$, $+0.0032$] & 42\% & [39\%, 43\%] & 0.0009/0.0008 \\
All compositions & 650 & $+0.0013$ & [$+0.0007$, $+0.0024$] & 46\% & [42\%, 47\%] & 0.0010/0.0010 \\
Unseen pairs & 416 & $+0.0003$ & [$-0.0006$, $+0.0015$] & 49\% & [44\%, 51\%] & 0.0014/0.0012 \\
Seen pairs & 234 & $+0.0034$ & [$+0.0030$, $+0.0044$] & 41\% & [38\%, 42\%] & 0.0006/0.0008 \\
Interpolation & 60 & $+0.0055$ & [$+0.0048$, $+0.0062$] & 2\% & [-3\%, 2\%] & 0.0015/0.0012 \\
\bottomrule
\end{tabular}
}
\end{table}

\begin{table}[H]
\centering
\caption{Confirmation experiment: mean curve error $M$ per monitor and set (seed mean $\pm$ seed SD). $^\dagger$Trained with condition labels.}
\label{S:tab:cont_arms_conf}
\resizebox{\textwidth}{!}{%
\setlength{\tabcolsep}{3pt}
\begin{tabular}{lccccc}
\toprule
Monitor & All (710) & Compositions (650) & Unseen pairs (416) & Seen pairs (234) & Interpolation (60) \\
\midrule
Functional decoder, 32 knots & 0.0767$\pm$0.0009 & 0.0796$\pm$0.0010 & 0.0810$\pm$0.0014 & 0.0772$\pm$0.0006 & 0.0451$\pm$0.0015 \\
Soft lookup$^\dagger$ & 0.0801$\pm$0.0008 & 0.0839$\pm$0.0010 & 0.0873$\pm$0.0012 & 0.0780$\pm$0.0008 & 0.0384$\pm$0.0012 \\
Hard lookup$^\dagger$ & 0.0900$\pm$0.0002 & 0.0937$\pm$0.0002 & 0.0960$\pm$0.0003 & 0.0896$\pm$0.0003 & 0.0498$\pm$0.0018 \\
Functional decoder, 16 knots & 0.0855$\pm$0.0012 & 0.0887$\pm$0.0014 & 0.0905$\pm$0.0018 & 0.0855$\pm$0.0007 & 0.0508$\pm$0.0023 \\
Mean behavior & 0.2064 & 0.2107 & 0.2036 & 0.2235 & 0.1596 \\
Brightness & 0.1969$\pm$0.0012 & 0.2055$\pm$0.0011 & 0.2068$\pm$0.0015 & 0.2033$\pm$0.0004 & 0.1042$\pm$0.0030 \\
Score only & 0.1402$\pm$0.0008 & 0.1430$\pm$0.0006 & 0.1395$\pm$0.0010 & 0.1492$\pm$0.0006 & 0.1094$\pm$0.0030 \\
\bottomrule
\end{tabular}
}
\end{table}

\begin{table}[H]
\centering
\caption{Confirmation experiment: $\Delta$ by quartile of the distance $d$ to the nearest training condition (650 compositions).}
\label{S:tab:cont_quart}
\small
\begin{tabular}{lcccc}
\toprule
Quartile of $d$ & $d$ range & Conditions & Mean $\Delta$ [basic 95\% CI] & Median $\Delta_c$ [basic 95\% CI] \\
\midrule
Q1 & 0.06--0.24 & 163 & $-0.0011$ [$-0.0018$, $-0.0003$] & $+0.0025$ [$+0.0017$, $+0.0037$] \\
Q2 & 0.24--0.34 & 162 & $-0.0018$ [$-0.0029$, $-0.0012$] & $+0.0043$ [$+0.0041$, $+0.0054$] \\
Q3 & 0.34--0.51 & 162 & $-0.0054$ [$-0.0063$, $-0.0041$] & $+0.0001$ [$-0.0015$, $+0.0017$] \\
Q4 & 0.51--0.96 & 163 & $-0.0091$ [$-0.0096$, $-0.0085$] & $-0.0038$ [$-0.0048$, $-0.0018$] \\
\bottomrule
\end{tabular}

\end{table}

\begin{table}[H]
\centering
\caption{Confirmation experiment: share of compositions whose per-condition error exceeds the soft lookup's 90th-percentile validation error (threshold fixed before test matching).}
\label{S:tab:cont_tail}
\small
\begin{tabular}{lccc}
\toprule
Monitor & All compositions & Unseen pairs & Seen pairs \\
\midrule
Functional decoder, 32 knots & 13\% & 14\% & 13\% \\
Soft lookup$^\dagger$ & 18\% & 19\% & 16\% \\
Hard lookup$^\dagger$ & 23\% & 24\% & 22\% \\
Functional decoder, 16 knots & 18\% & 19\% & 18\% \\
Mean behavior & 65\% & 62\% & 70\% \\
Brightness & 64\% & 61\% & 70\% \\
Score only & 68\% & 64\% & 74\% \\
\bottomrule
\end{tabular}

\end{table}

\subsection{Verification and Changes During Execution}
Every check stopped the pipeline on failure; failures are documented, not repeated silently. The generator reproduced 3{,}500 of 3{,}500 stored deterministic degradations bit for bit, and stochastic degradations were bit-identical across processes. Inference and training were reproducible bit for bit, including interruption and resumption of training. The frozen monitors reproduced the initial estimates within $10^{-16}$. Pooled operating curves agreed with the main study's stored detections within a mean curve distance of 0.00025 (tolerance 0.002); a pre-specified detection-level matching criterion (99\%) was not met (97.4\%), owing to batched versus single-image inference and a half-pixel coordinate offset in the stored detections, so equivalence was established at the curve level. All monitors and targets within these experiments use one uniform pipeline. In the initial experiment, the monitors were trained with fp16 mixed precision (autocast with gradient scaling) and TF32 matrix multiplication. Detector inference (feature extraction and detections) ran with FP32 matrix multiplication for the training and validation conditions but, because of a setting left over from runtime measurements earlier in the same process, with TF32 matrix multiplication for the test conditions. A re-inference of all 142 test conditions with FP32 matrix multiplication, with recomputed targets, changed the decoder--lookup differences and their percentile and basic interval limits by less than $10^{-5}$ (at most $7.1\times10^{-6}$). In the confirmation experiment, which trained no monitors, every detector-inference job set and verified FP32 matrix multiplication before running. Monitor-head predictions in both experiments used TF32 matrix multiplication without autocast. Convolutions used the PyTorch default, which permits TF32 in cuDNN. Changes made during execution, each fixed in writing before the affected step and without access to test outcomes: the spatter-water range was moved to $[1.5,3]$ after a generator agreement check failed at a near-identity cell; the group-size candidates were extended; the monotonicity check was repeated with common random numbers; measured runtimes replaced a projection formula and the organizational runtime limit was raised; detector equivalence was assessed at the curve level; and the reference of one inference check was moved to a training condition because of the precision difference.

\subsection{Limitations}
\begin{itemize}
\item \textbf{Mean, not typical case.} In the typical composition the lookup is slightly more accurate (median $\Delta_c=+0.0013$; decoder better in 46\%). Without the 10\% most favorable conditions, the mean difference is $+0.0009$. The advantage comes from an asymmetric tail: the decoder is better by more than 0.02 in 14.8\% of compositions and worse by more than 0.02 in 2.9\%, and it exceeds the lookup's validation error threshold less often (13.4\% versus 18.2\%).
\item \textbf{Where the advantage arises.} It is concentrated in compositions of pairs never combined in training ($\Delta=-0.0063$); for seen pairs at new strength combinations it is small ($-0.0008$, and the typical condition favors the lookup). Heterogeneity across pairs is large: the decoder is better for 14 of 26 unseen pairs, with pair means from $-0.046$ to $+0.008$. The association with distance is real but weak ($\rho=-0.18$) and uses an assumed scaling of the severity coordinates.
\item \textbf{Interpolation favors the lookup} in both experiments ($\Delta=+0.0067$), in all 15 types in the confirmation experiment.
\item \textbf{Same pairs and base images.} The confirmation experiment draws fresh strengths and combinations, but uses the same 26 unseen pairs and the same 1{,}650 test base images; the base images were never used for training or model selection. The result generalizes to new strengths of these pairs, not to arbitrary pairs or images.
\item \textbf{One training run} (5 seeds for decoder and lookup, 3 otherwise); seed SDs describe these models, not a new training campaign.
\item \textbf{Representation.} The advantage was shown for the 32-knot decoder. The 16-knot decoder of the main study is worse than the lookup ($\Delta=+0.0054$ over all initial test conditions, $+0.0047$ on the confirmation compositions).
\item \textbf{Interval construction.} Recomputing target curves in each resample shifts and skews the replicate distribution; the initial endpoint therefore depends on the interval type, whereas all confirmation results hold for every variant.
\item \textbf{Scope.} One detector, one class, one image source; synthetic degradations on the common-corruptions scale; 15 types, pairs in one fixed order; one store density (10 strengths per type, 2 combinations per training pair), which shifts the balance between interpolation and composition. Operational differences are small (execution $F_1$ $+0.0012$ in favor of the decoder in the confirmation experiment), and the oracle replicate shows large headroom ($M=0.015$ versus about 0.069).
\end{itemize}

\clearpage
\section{Natural-Condition Studies}
\label{S:sec:natural_supp}
This section documents the natural-condition studies referred to in the Discussion of the main paper (``Coverage and generality''): data preparation, regimes, protocol, and full results. They address whether DOBI applies beyond synthetic probes, to coarse natural conditions given by dataset metadata. All studies use the frozen COCO YOLOv10-Nano of the main study without any detector change.

\paragraph{Studies and status.}
\begin{itemize}
\item \textbf{Vehicle study (BDD100K):} phase~1 compared weather regimes (exploratory); phase~1b compared scene $\times$ time-of-day regimes; a rule fixed before training chose the regime family for the monitor study; phase~2 trained and tested monitors with pre-specified hypotheses.
\item \textbf{Person study (BDD100K):} pre-specified confirmation of the vehicle study for the class of the main paper, with its regimes fixed in advance; written before any person count, test-target computation or training run.
\item \textbf{Reporting commitment:} the person-study protocol requires both studies to be reported together.
\end{itemize}

\subsection{BDD100K Data Preparation}
\paragraph{Source.} BDD100K (Yu et al., CVPR 2020) (100K images, detection labels in the 2018 release format). Label files: \texttt{bdd100k\_labels\_images\_train.json} (69{,}863 entries) and \texttt{bdd100k\_labels\_images\_val.json} (10{,}000 entries). The test split has no public labels and is not used. All 79{,}863 labelled images are $1280\times720$. Images were used locally only; none are shown.

\paragraph{Labels and class mapping.} \Cref{S:tab:bdd_t4} gives the mapping. \emph{Vehicle} (primary, ``V-narrow''): car, truck, bus on ground-truth and detector side; train, motorcycle and bicycle ground truth are ignore regions. A broader variant with all vehicle classes shows the same pattern. \emph{Person}: ground truth person and rider merged; detector class person. The 2018 labels have no crowd flag. An unmatched detection lying more than 50\% inside the union of ignore boxes is removed. No size filter is applied.

\paragraph{Metadata and regimes.} Each image carries one value for weather, time of day and scene (\Cref{S:tab:bdd_t1,S:tab:bdd_t2}); there are no severity levels. Images with an undefined value in a regime-defining attribute belong to no regime. The six regimes are \{city street, highway, residential\} $\times$ \{daytime, night\}, independent of weather; dawn/dusk and the scenes parking lot, tunnel and gas stations are not used. Qualification for the vehicle study: $\geq$\,200 images and $\geq$\,1{,}000 vehicles per regime and split (all six qualify). For the person study, a test regime needs $\geq$\,200 images and $\geq$\,500 persons; city-day, city-night and residential-day qualify, highway-day, highway-night and residential-night are descriptive only but remain in training.

\paragraph{Splits and independence.} Test is the official validation split; training uses the official train split. Each image is a key frame of a different video, but the first block of the image name groups up to 48 images; we treat this prefix as the independence unit. Only one prefix is shared between official train and val; the three training images with that prefix are excluded. The validation part for checkpoint selection consists of about 15\% of the training prefixes per regime, drawn with a fixed seed. No prefix occurs in more than one part (\Cref{S:tab:bdd_t5}). Sizes for the scene regimes: training 50{,}389, validation 13{,}568, test 9{,}087 images (\Cref{S:tab:bdd_t3}).

\paragraph{Detector input and evaluation.} Images are resized to a long side of 640 and letterboxed exactly as in the main study; batches are padded to full size, which makes features and detections bitwise reproducible. Matching, score floor (0.001), threshold grid and curve definitions are those of the main study. FROC and PR are the primary curves; ROC is reported only, because the main-study FPR definition is dominated by low-score false positives on these images. The 300-detection cap is reached by 43.8\% of images but truncates only scores below 0.008. Size bands are measured at the detector input (box height $\times$ 0.5).

\paragraph{Monitor protocol and statistics.} Monitors follow the main protocol (PCHIP-16 primary, PCHIP-32 secondary, three seeds). Training samples regimes uniformly per epoch. Comparators: the mean-behavior prior; a brightness baseline (same decoder on luminance statistics of the image); the soft lookup (regime classifier on the same features, probability-weighted mixture of stored curves, precision recomputed from mixed TPR and FPPI); and the non-deployable regime oracle. Primary metric $M=\tfrac12(\mathrm{MAD}_{\rm FROC}+\mathrm{MAD}_{\rm PR})$, averaged over qualified test regimes; the natural-distribution average over all test images is reported alongside. Confidence intervals are paired percentile bootstraps over test prefixes (2{,}000 resamples, seed 12345, stratified by regime) in which the regime target curves are recomputed in every resample. Hypotheses: H1, DOBI better than the mean-behavior prior; H2, DOBI better than the brightness baseline; supported if the upper CI bound is below zero.

\begin{table}[H]
\centering
\caption{BDD100K (2018 labels): images per split by weather and time of day.}
\label{S:tab:bdd_t1}
\small
\begin{tabular}{llrr}
\toprule
Weather & Time of day & Train & Val \\
\midrule
clear & night & 22,884 & 3,274 \\
clear & daytime & 12,454 & 1,764 \\
overcast & daytime & 7,551 & 1,039 \\
undefined & daytime & 7,029 & 994 \\
partly cloudy & daytime & 4,262 & 638 \\
snowy & daytime & 2,862 & 422 \\
rainy & daytime & 2,522 & 396 \\
snowy & night & 2,249 & 273 \\
rainy & night & 2,208 & 286 \\
clear & dawn/dusk & 2,004 & 307 \\
overcast & dawn/dusk & 1,147 & 182 \\
partly cloudy & dawn/dusk & 570 & 95 \\
undefined & dawn/dusk & 527 & 64 \\
undefined & night & 442 & 66 \\
snowy & dawn/dusk & 436 & 74 \\
rainy & dawn/dusk & 328 & 55 \\
undefined & undefined & 121 & 33 \\
overcast & night & 72 & 18 \\
foggy & night & 67 & 7 \\
partly cloudy & night & 49 & 5 \\
foggy & daytime & 48 & 5 \\
foggy & dawn/dusk & 15 & 1 \\
rainy & undefined & 12 & 1 \\
snowy & undefined & 2 & 0 \\
clear & undefined & 2 & 1 \\
total &  & 69,863 & 10,000 \\
\bottomrule
\end{tabular}

\end{table}

\begin{table}[H]
\centering
\caption{BDD100K: images per split by scene and time of day.}
\label{S:tab:bdd_t2}
\small
\begin{tabular}{llrr}
\toprule
Scene & Time of day & Train & Val \\
\midrule
city street & daytime & 21,811 & 3,067 \\
city street & night & 18,748 & 2,582 \\
highway & daytime & 8,905 & 1,290 \\
highway & night & 7,025 & 1,007 \\
residential & daytime & 5,658 & 862 \\
city street & dawn/dusk & 2,950 & 461 \\
residential & night & 1,813 & 279 \\
highway & dawn/dusk & 1,439 & 200 \\
residential & dawn/dusk & 599 & 111 \\
undefined & night & 264 & 40 \\
parking lot & daytime & 228 & 28 \\
parking lot & night & 94 & 15 \\
undefined & daytime & 77 & 7 \\
tunnel & undefined & 76 & 22 \\
tunnel & daytime & 32 & 2 \\
parking lot & dawn/dusk & 30 & 2 \\
parking lot & undefined & 25 & 4 \\
tunnel & night & 20 & 3 \\
gas stations & daytime & 17 & 2 \\
undefined & undefined & 15 & 4 \\
highway & undefined & 10 & 2 \\
city street & undefined & 7 & 2 \\
gas stations & night & 7 & 3 \\
undefined & dawn/dusk & 5 & 2 \\
residential & undefined & 4 & 1 \\
gas stations & dawn/dusk & 3 & 2 \\
tunnel & dawn/dusk & 1 & 0 \\
total &  & 69,863 & 10,000 \\
\bottomrule
\end{tabular}

\end{table}

\begin{table}[H]
\centering
\caption{Scene $\times$ time-of-day regimes per part: images, name prefixes, persons (person + rider), riders, vehicles (car + truck + bus), and the percentage of ground-truth boxes per height band at the detector input (P: persons, V: vehicles).}
\label{S:tab:bdd_t3}
\resizebox{\textwidth}{!}{%
\begin{tabular}{llrrrrrrrrrrr}
\toprule
Regime & Part & Images & Prefixes & Persons & Riders & Vehicles & P$<$32 & P 32--64 & P$\geq$64 & V$<$32 & V 32--64 & V$\geq$64 \\
\midrule
city day & training & 17,467 & 13,103 & 50,147 & 2,240 & 207,394 & 63 & 27 & 9 & 70 & 17 & 13 \\
city night & training & 14,858 & 10,115 & 15,005 & 617 & 146,641 & 57 & 32 & 11 & 70 & 18 & 12 \\
highway day & training & 6,978 & 5,911 & 2,174 & 226 & 76,878 & 73 & 20 & 7 & 82 & 11 & 7 \\
highway night & training & 5,393 & 4,535 & 479 & 58 & 41,870 & 63 & 27 & 10 & 80 & 13 & 7 \\
residential day & training & 4,372 & 3,770 & 2,822 & 197 & 55,522 & 63 & 27 & 10 & 72 & 17 & 12 \\
residential night & training & 1,321 & 1,222 & 249 & 27 & 15,405 & 45 & 33 & 21 & 63 & 21 & 16 \\
city day & validation & 4,342 & 3,049 & 12,672 & 534 & 50,985 & 63 & 28 & 10 & 70 & 17 & 13 \\
city night & validation & 3,890 & 2,380 & 3,737 & 145 & 37,395 & 56 & 33 & 12 & 70 & 18 & 12 \\
highway day & validation & 1,926 & 1,581 & 615 & 52 & 21,112 & 74 & 21 & 6 & 81 & 12 & 7 \\
highway night & validation & 1,632 & 1,280 & 151 & 14 & 12,468 & 61 & 32 & 7 & 79 & 13 & 8 \\
residential day & validation & 1,286 & 1,098 & 780 & 74 & 16,193 & 62 & 27 & 12 & 72 & 17 & 11 \\
residential night & validation & 492 & 429 & 78 & 7 & 5,726 & 45 & 36 & 19 & 64 & 21 & 15 \\
city day & test & 3,067 & 2,248 & 8,817 & 394 & 36,683 & 63 & 28 & 9 & 71 & 17 & 13 \\
city night & test & 2,582 & 1,738 & 2,813 & 110 & 25,568 & 57 & 32 & 11 & 70 & 17 & 12 \\
highway day & test & 1,290 & 1,076 & 375 & 38 & 13,774 & 74 & 18 & 7 & 81 & 12 & 7 \\
highway night & test & 1,007 & 802 & 94 & 11 & 7,732 & 67 & 23 & 10 & 80 & 13 & 7 \\
residential day & test & 862 & 745 & 612 & 46 & 11,158 & 63 & 28 & 9 & 72 & 16 & 11 \\
residential night & test & 279 & 252 & 55 & 3 & 3,189 & 35 & 38 & 27 & 66 & 21 & 13 \\
\bottomrule
\end{tabular}
}
\end{table}

\begin{table}[H]
\centering
\caption{Class mapping. V-narrow is primary for vehicles.}
\label{S:tab:bdd_t4}
\resizebox{\textwidth}{!}{%
\begin{tabular}{llll}
\toprule
Side & Class (label name) & Group & Use \\
\midrule
GT & car & vehicle (V-narrow, V-broad) & evaluated \\
GT & truck & vehicle (V-narrow, V-broad) & evaluated \\
GT & bus & vehicle (V-narrow, V-broad) & evaluated \\
GT & train & vehicle (V-broad); ignore region (V-narrow) & evaluated / ignore \\
GT & motor & motorcycle: vehicle (V-broad); ignore region (V-narrow) & evaluated / ignore \\
GT & bike & bicycle: vehicle (V-broad); ignore region (V-narrow) & evaluated / ignore \\
GT & person & person & evaluated (person study) \\
GT & rider & person & evaluated (person study) \\
GT & traffic light, traffic sign, lane, drivable area & - & not used \\
detector (COCO) & car, truck, bus & vehicle (V-narrow, V-broad) & kept \\
detector (COCO) & bicycle, motorcycle, train, boat, airplane & vehicle (V-broad only; COCO\_LABELS) & kept (V-broad) / discarded (V-narrow) \\
detector (COCO) & person & person & kept (person study) \\
detector (COCO) & all other 71 classes & - & discarded \\
\bottomrule
\end{tabular}
}
\end{table}

\begin{table}[H]
\centering
\caption{Exclusions and leakage checks.}
\label{S:tab:bdd_t5}
\small
\begin{tabular}{p{0.5\textwidth}rp{0.3\textwidth}}
\toprule
Item & Count & Handling / result \\
\midrule
train images on disk without a label entry & 137 & not used (no labels) \\
label entries without image (train / val) & 0 & - \\
image-name prefixes shared between official train and val & 1 & train images with that prefix excluded \\
train images excluded because their prefix occurs in the test set (scene regimes) & 3 & prefix fe189115 \\
prefixes in more than one part (vehicle study) & 0 & leakage check passed \\
prefixes in more than one part (person study) & 0 & leakage check passed \\
labelled images outside the six scene x time-of-day regimes (train+val) & 6,816 & dawn/dusk, undefined, other scenes \\
images reaching the 300-detection cap (all classes) & 35,002 & 43.8 \%; truncated scores $<$ 0.0080 \\
Person-study test regimes below the size rule ($\geq$ 200 images and $\geq$ 500 persons) & 3 & descriptive only: highway day, highway night, residential night \\
\bottomrule
\end{tabular}

\end{table}

\subsection{Choice of Regimes (Vehicle Study)}
\emph{Weather regimes (phase~1, exploratory).} With daytime clear weather as reference, night regimes (clear, rainy, snowy night) lie at behavior distances of 1.77--2.36 from it (Experiment-B distance, vehicles, pooled), against a largest same-regime train--validation distance of 0.44; this replicates on the validation split and holds in every size band. Daytime overcast, rain and snow lie at 0.41--0.50, i.e.\ at the noise level. Foggy images are too few (130/13). Adapting the threshold per weather regime gains at most $+0.014$ $F_1$ (clear night) pooled; larger gains appear only for large objects at night (\Cref{S:tab:nat_weather}). The exploratory go criterion for a monitor study (two regimes beyond noise and a threshold gain $\geq$\,0.02 with CI above zero) was met only in the $\geq$\,64\,px view, carried by clear night.
\emph{Scene $\times$ time-of-day regimes (phase~1b).} Daytime scene regimes differ clearly (city--highway 1.14, highway--residential 1.19, city--residential 0.42 against a noise maximum of 0.29). By the rule fixed before monitor training, the monitor studies therefore use these regimes (\Cref{S:tab:nat_scenes}).

\begin{table}[H]
\centering
\caption{Weather regimes (vehicle study, phase~1, exploratory). Left: behavior distance to clear day on training data, pooled and by box height at the detector input; the noise threshold is the largest distance between training and validation curves of the same regime. Right: $F_1$ gain of a regime-specific threshold over the clear-day threshold. $^\ast$Foggy: too few images (130/13), descriptive.}
\label{S:tab:nat_weather}
\small
\begin{tabular}{lcccccc}
\toprule
 & \multicolumn{4}{c}{Distance to clear day (by box height)} & \multicolumn{2}{c}{Threshold gain $\Delta F_1$} \\
\cmidrule(lr){2-5}\cmidrule(lr){6-7}
Regime & pooled & $<$32 & 32--64 & $\geq$64 & pooled & $\geq$64 \\
\midrule
overcast, day & 0.41 & 0.51 & 0.43 & 0.56 & $-0.0008$ & $-0.0002$ \\
rainy, day & 0.50 & 0.54 & 0.54 & 0.93 & $-0.0014$ & $+0.0055$ \\
snowy, day & 0.42 & 0.42 & 0.50 & 0.95 & $-0.0034$ & $+0.0048$ \\
dawn/dusk & 0.43 & 0.22 & 0.55 & 1.00 & $+0.0002$ & $-0.0001$ \\
clear, night & 1.88 & 1.83 & 1.75 & 1.91 & $+0.0141$ & $+0.0315$ \\
rainy, night & 2.36 & 2.43 & 2.43 & 2.11 & $-0.0010$ & $+0.0252$ \\
snowy, night & 1.77 & 1.85 & 1.79 & 1.90 & $+0.0051$ & $+0.0176$ \\
foggy$^\ast$ & 1.53 & 1.03 & 1.80 & 1.44 & $+0.0046$ & $+0.1077$ \\
\midrule
Noise threshold & 0.44 & 0.61 & 0.51 & 0.82 & -- & -- \\
\bottomrule
\end{tabular}

\end{table}

\begin{table}[H]
\centering
\caption{Daytime scene regimes (vehicle study, phase~1b): pairwise behavior distances. All pairs exceed the noise threshold, so the predefined rule selected scene $\times$ time-of-day regimes.}
\label{S:tab:nat_scenes}
\small
\begin{tabular}{lcc}
\toprule
Daytime scene pair & Distance (train) & Distance (val) \\
\midrule
city day -- highway day & 1.14 & 1.17 \\
city day -- residential day & 0.42 & 0.44 \\
highway day -- residential day & 1.19 & 1.21 \\
\midrule
Noise threshold (max.\ same-regime train--val) & \multicolumn{2}{c}{0.29} \\
\bottomrule
\end{tabular}

\end{table}
\emph{Data-use caveat.} The rule's decision statistic used training data only, but its noise threshold and qualification counts also used the official validation split, i.e.\ the later test set; phases~1 and~1b also computed descriptive vehicle curves on it. For the person study, the regimes were taken over unchanged and the test split was not touched before its protocol was fixed.

\subsection{Results}
\begin{table}[H]
\centering
\caption{Natural-condition monitors on BDD100K. $M$ is the curve error (lower is better). ``Regime mean'' averages over the test regimes with enough data for a reliable target curve: all 6 for vehicles; for persons the 3 regimes with $\geq$\,200 images and $\geq$\,500 persons (pre-specified). The 95\% CI resamples test images and target curves. ``Natural'' weights regimes by their number of test images; ``day'' and ``night'' average the corresponding regimes (person night: city-night only). $^\dagger$Trained with regime labels. $^\ddagger$Knows the true regime; not deployable.}
\label{S:tab:nat_arms}
\resizebox{\textwidth}{!}{%
\begin{tabular}{llccccc}
\toprule
Study & Arm & $M$ regime mean [95\% CI] & seed SD & $M$ natural & $M$ day & $M$ night \\
\midrule
Vehicles & Mean behavior & 0.0684 [0.0643, 0.0727] & 0.0000 & 0.0726 & 0.0698 & 0.0670 \\
 & Brightness & 0.0366 [0.0330, 0.0412] & 0.0002 & 0.0294 & 0.0359 & 0.0374 \\
 & DOBI (PCHIP-16) & 0.0308 [0.0291, 0.0342] & 0.0002 & 0.0287 & 0.0313 & 0.0303 \\
 & Soft lookup$^\dagger$ & 0.0237 [0.0214, 0.0281] & 0.0015 & 0.0199 & 0.0217 & 0.0257 \\
 & Regime oracle$^\ddagger$ & 0.0067 [0.0054, 0.0118] & 0.0000 & 0.0054 & 0.0057 & 0.0077 \\
 & DOBI (PCHIP-32) & 0.0271 [0.0253, 0.0306] & 0.0001 & 0.0249 & 0.0267 & 0.0274 \\
\midrule
Persons & Mean behavior & 0.0587 [0.0437, 0.0761] & 0.0000 & 0.0549 & 0.0775 & 0.0213 \\
 & Brightness & 0.0409 [0.0366, 0.0549] & 0.0011 & 0.0418 & 0.0417 & 0.0394 \\
 & DOBI (PCHIP-16) & 0.0362 [0.0332, 0.0500] & 0.0003 & 0.0355 & 0.0361 & 0.0365 \\
 & Soft lookup$^\dagger$ & 0.0304 [0.0264, 0.0447] & 0.0003 & 0.0266 & 0.0343 & 0.0225 \\
 & Regime oracle$^\ddagger$ & 0.0116 [0.0094, 0.0308] & 0.0000 & 0.0098 & 0.0123 & 0.0101 \\
 & DOBI (PCHIP-32) & 0.0360 [0.0339, 0.0494] & 0.0002 & 0.0368 & 0.0346 & 0.0388 \\
\bottomrule
\end{tabular}
}
\end{table}

\begin{table}[H]
\centering
\caption{Paired contrasts ($\Delta$ = first minus second arm; negative $\Delta M$ and positive $\Delta F_1$ favor the first). H1 and H2 are the pre-specified primary contrasts.}
\label{S:tab:nat_contrasts}
\small
\begin{tabular}{lcc}
\toprule
Contrast & Vehicles & Persons \\
\midrule
H1: DOBI $-$ mean (binding) & $-0.0376$ [$-0.0423$, $-0.0314$] & $-0.0225$ [$-0.0270$, $-0.0067$] \\
H2: DOBI $-$ brightness (binding) & $-0.0059$ [$-0.0073$, $-0.0037$] & $-0.0047$ [$-0.0057$, $-0.0023$] \\
DOBI $-$ mean, natural distribution & $-0.0439$ [$-0.0470$, $-0.0398$] & $-0.0194$ [$-0.0234$, $-0.0123$] \\
DOBI $-$ brightness, natural distribution & $-0.0006$ [$-0.0020$, $+0.0008$] & $-0.0063$ [$-0.0073$, $-0.0038$] \\
DOBI $-$ brightness, day & $-0.0047$ [$-0.0067$, $-0.0020$] & $-0.0056$ [$-0.0072$, $-0.0023$] \\
DOBI $-$ brightness, night & $-0.0070$ [$-0.0092$, $-0.0034$] & $-0.0029$ [$-0.0041$, $+0.0001$] \\
DOBI $-$ mean, night & $-0.0366$ [$-0.0449$, $-0.0269$] & $+0.0152$ [$+0.0130$, $+0.0177$] \\
DOBI $-$ soft lookup & $+0.0071$ [$+0.0058$, $+0.0079$] & $+0.0059$ [$+0.0030$, $+0.0085$] \\
Executed $F_1$: DOBI $-$ fixed $\tau$ & $+0.0050$ [$+0.0034$, $+0.0066$] & $+0.0063$ [$+0.0023$, $+0.0102$] \\
Executed $F_1$: soft lookup $-$ fixed $\tau$ & $+0.0050$ [$+0.0033$, $+0.0067$] & $+0.0011$ [$-0.0005$, $+0.0026$] \\
\bottomrule
\end{tabular}

\end{table}

\begin{table}[H]
\centering
\caption{Curve error $M$ per test regime (seed means; lower is better). $^\ast$Person regimes with too few test persons for a reliable target curve (fewer than 500; pre-specified rule): reported for completeness, not used in the averages. Their high oracle values show how noisy such targets are.}
\label{S:tab:nat_regimes}
\resizebox{\textwidth}{!}{%
\begin{tabular}{llcccccc}
\toprule
Study & Arm & city day & city night & highway day & highway night & resid.\ day & resid.\ night \\
\midrule
Vehicles & Mean behavior & 0.087 & 0.061 & 0.024 & 0.111 & 0.099 & 0.029 \\
 & Brightness & 0.020 & 0.017 & 0.057 & 0.046 & 0.030 & 0.048 \\
 & DOBI (PCHIP-16) & 0.026 & 0.024 & 0.041 & 0.034 & 0.027 & 0.033 \\
 & Soft lookup$^\dagger$ & 0.017 & 0.016 & 0.025 & 0.025 & 0.024 & 0.036 \\
 & Regime oracle$^\ddagger$ & 0.004 & 0.004 & 0.007 & 0.008 & 0.006 & 0.011 \\
\midrule
Persons & Mean behavior & 0.077 & 0.021 & 0.040$^\ast$ & 0.168$^\ast$ & 0.078 & 0.066$^\ast$ \\
 & Brightness & 0.045 & 0.039 & 0.078$^\ast$ & 0.145$^\ast$ & 0.039 & 0.086$^\ast$ \\
 & DOBI (PCHIP-16) & 0.034 & 0.036 & 0.061$^\ast$ & 0.125$^\ast$ & 0.038 & 0.071$^\ast$ \\
 & Soft lookup$^\dagger$ & 0.025 & 0.023 & 0.059$^\ast$ & 0.131$^\ast$ & 0.043 & 0.071$^\ast$ \\
 & Regime oracle$^\ddagger$ & 0.008 & 0.010 & 0.030$^\ast$ & 0.086$^\ast$ & 0.017 & 0.058$^\ast$ \\
\bottomrule
\end{tabular}
}
\end{table}

\begin{table}[H]
\centering
\caption{Vehicles: for each true test regime (rows, with image counts), the percentage of test images whose predicted curve is closest to each training-regime curve (DOBI-16, seed mean). Diagonal in bold.}
\label{S:tab:nat_nearest_v}
\small
\begin{tabular}{lcccccc}
\toprule
True $\backslash$ nearest & city day & city night & highway day & highway night & resid. day & resid. night \\
\midrule
city day (3,067) & \textbf{69} & 0 & 11 & 0 & 20 & 1 \\
city night (2,582) & 0 & \textbf{69} & 3 & 9 & 0 & 19 \\
highway day (1,290) & 37 & 2 & \textbf{60} & 0 & 1 & 1 \\
highway night (1,007) & 0 & 45 & 1 & \textbf{50} & 0 & 4 \\
resid. day (862) & 62 & 1 & 4 & 0 & \textbf{32} & 1 \\
resid. night (279) & 1 & 53 & 3 & 3 & 0 & \textbf{40} \\
\bottomrule
\end{tabular}

\end{table}

\begin{table}[H]
\centering
\caption{Persons: as \Cref{S:tab:nat_nearest_v}. Highway day, highway night and residential night are descriptive regimes (too few test persons).}
\label{S:tab:nat_nearest_p}
\small
\begin{tabular}{lcccccc}
\toprule
True $\backslash$ nearest & city day & city night & highway day & highway night & resid. day & resid. night \\
\midrule
city day (3,067) & \textbf{58} & 13 & 2 & 0 & 26 & 0 \\
city night (2,582) & 3 & \textbf{39} & 46 & 9 & 0 & 2 \\
highway day (1,290) & 8 & 37 & \textbf{36} & 0 & 18 & 1 \\
highway night (1,007) & 0 & 7 & 43 & \textbf{49} & 0 & 1 \\
resid. day (862) & 24 & 9 & 2 & 0 & \textbf{65} & 0 \\
resid. night (279) & 1 & 14 & 65 & 3 & 2 & \textbf{15} \\
\bottomrule
\end{tabular}

\end{table}

\paragraph{Findings.}
\begin{itemize}
\item \textbf{H1 and H2 are supported for vehicles and persons} (\Cref{S:tab:nat_contrasts}). For persons this also holds in the natural-distribution average; for vehicles, DOBI versus brightness is undecided in that average, because brightness is better in city scenes, which make up most test images, and worse on highway and residential scenes.
\item \textbf{Beyond day versus night.} For vehicles, DOBI beats brightness within day and within night. For persons, it beats brightness within day; at night (city only), the comparison is undecided and the mean-behavior prior, which happens to lie close to the city-night curve, is better than DOBI.
\item \textbf{Regime structure.} Matching each prediction to its nearest training curve separates day and night for vehicles almost perfectly (at most 3\% cross-assignments) but not for persons (46\% of city-night images are closest to the highway-day curve); scene assignments are correct for 32--69\% (vehicles) and 39--65\% (persons, qualified regimes) of images (\Cref{S:tab:nat_nearest_v,S:tab:nat_nearest_p}).
\item \textbf{Regime-supervised lookup.} As in the follow-up studies of \Cref{S:sec:followup}, the soft lookup is the best deployable arm (DOBI minus soft lookup $+0.0071$ vehicles, $+0.0059$ persons), and the oracle shows large remaining headroom.
\item \textbf{Operating decisions.} Executing the predicted threshold gains $+0.005$ (vehicles) and $+0.006$ (persons) $F_1$ over a fixed threshold; for persons the soft lookup gains nothing measurable. Within size bands, no person regime shows a threshold gain.
\item \textbf{Composition.} Object size and density differ between scenes (\Cref{S:tab:bdd_t3}); part of the regime signal therefore reflects scene composition rather than image quality.
\end{itemize}

\paragraph{Limitations of the natural-condition studies.} Metadata are coarse per-image categories without intensity; the regimes were chosen after exploratory analysis on vehicles (with the data-use caveat above); the person study has three qualified regimes; scene composition is a confounder; monitors are trained and tested within one domain, so cross-domain transfer is not shown.

\section{Reproducibility}
\label{S:sec:repro}
Main models are trained for 60 epochs with three seeds; ablations and held-out runs use seed 0. The execution analysis uses all 85,800 test variants and 2,685,600 stored detections. A fixed threshold executed image by image reproduces the stored regime sweeps exactly, and the regime oracle reproduces the stored optimal points to numerical precision. Bootstrap intervals use 2,000 paired resamples of base images with seed 12345. No model was retrained for the original Nano execution analysis. The separately trained Medium-detector monitors are documented in \Cref{S:sec:supp_medium}; their stored-detector execution and score-floor check use independently verified detector-specific outputs. Code, configurations, split definitions, and the time-stamped study protocols are available from the authors.

\end{document}